\documentclass[journal]{IEEEtran}
\usepackage{amsmath,amssymb,amsfonts}
\usepackage{cases}
\usepackage{color,xcolor}
\usepackage{graphicx}
\usepackage{subfigure}
\usepackage{algorithm}
\usepackage{algorithmic}
\usepackage{epstopdf}
\usepackage{multirow}
\usepackage{rotating}
\usepackage{booktabs}
\usepackage{tikz}
\usepackage{acronym}
\usepackage{epstopdf}
\usepackage{pifont} 
\usepackage{newtxmath}
\usepackage[demo,
            export]{adjustbox}  
\usepackage{tabularray}
\usepackage{makecell}

\usepackage{colortbl}

\usepackage[colorlinks,
linkcolor=blue,
anchorcolor=blue,
citecolor=blue
]{hyperref}
\usepackage{amssymb}
\usepackage{array} 

\definecolor{tableTitleColor}{rgb}{0.8,0.8,0.8}

\newtheorem{Theorem}{Theorem}[section]
\newtheorem{Lemma}[Theorem]{Lemma}

\newtheorem{Remark}[Theorem]{Remark}
\newtheorem{Definition}[Theorem]{Definition}

\begin{document}

\title{Data-Adaptive Transformed Bilateral Tensor Low-Rank Representation for Clustering}

\author{Hui Chen, \IEEEmembership{Member, IEEE}, Xinjie Wang, Xianchao Xiu, \IEEEmembership{Member, IEEE}, \\and Wanquan Liu, \IEEEmembership{Senior Member, IEEE}

\thanks{This work was supported in part by the Natural Science Foundation of Shanghai under Grant 24ZR1425700, the National Natural Science Foundation of China under Grant 12371306, and the Project of the State Administration of Foreign Experts under Grant H20240974. (\textit{Corresponding author: Xianchao Xiu.})}
\thanks{H. Chen and X. Wang are with the School of Automation Engineering, Shanghai University of Electric Power, Shanghai 200090, China (e-mail: chenhui@shiep.edu.cn; 1840403867@mail.shiep.edu.cn).}
\thanks{X. Xiu is with the School of Mechatronic Engineering and Automation, Shanghai University, Shanghai 200444, China (e-mail: xcxiu@shu.edu.cn).}
\thanks{W. Liu is with the School of Intelligent Systems Engineering, Sun Yat-sen University, Guangzhou 510275, China (e-mail: liuwq63@mail.sysu.edu.cn).}
}

\maketitle
\begin{abstract}
Tensor low-rank representation (TLRR) has demonstrated significant success in image clustering. However, most existing methods rely on fixed transformations and suffer from poor robustness to noise. In this paper, we propose a novel transformed bilateral tensor low-rank representation model called TBTLRR, which introduces a data-adaptive tensor nuclear norm by learning arbitrary unitary transforms, allowing for more effective capture of global correlations. In addition, by leveraging the bilateral structure of latent tensor data, TBTLRR is able to exploit local correlations between image samples and features. Furthermore, TBTLRR integrates the $\ell_{1/2}$-norm and Frobenius norm regularization terms for better dealing with complex noise in real-world scenarios. To solve the proposed nonconvex model, we develop an efficient optimization algorithm inspired by the alternating direction method of multipliers (ADMM) and provide theoretical convergence. Extensive experiments validate its superiority over the state-of-the-art methods in  clustering. The code will be available at \url{https://github.com/xianchaoxiu/TBTLRR}.
\end{abstract}


\section{Introduction}\label{Introduction}

\IEEEPARstart{H}{igh}-dimensional data not only increases computational complexity but also degrades the performance of image processing and video analysis \cite{ray2021various}. Extensive studies have shown that such data often resides in low-dimensional subspaces and can be represented as a union of multiple unknown subspaces \cite{wright2022high}. Therefore, identifying low-dimensional subspaces that effectively capture the essential characteristics of the data is a critical issue \cite{jia2023semi,luo2024low}. 
Over the past few decades, low-rank representation (LRR) \cite{liu2012robust} has emerged as a representative subspace dimensionality reduction method, with broad applications in image denoising, face recognition, and hyperspectral analysis, see, e.g.,  \cite{xue2024tensor,li2024adaptive,wei2024learning,shen2025dual}.

The core of LRR is to decompose the observed data into a low-rank component that captures global correlations and a residual term that describes noise or outliers, thereby enabling accurate characterization of the global subspace structure \cite{zhou2014low}. However, when the number of available samples is insufficient, the performance of LRR will degrade significantly \cite{liu2010robust}. To address this issue, Liu et al. \cite{liu2011latent} proposed latent LRR (LLRR), which introduces latent variables to better characterize the intrinsic structure. However, the nuclear norm minimization in LLRR treats all singular values equally, which may result in a rank estimate that deviates from the actual rank. To this end, Liu et al. \cite{liu2023latLRR} developed an iterative reweighted Frobenius norm regularized LLRR (IRFLLRR) method that assigns adaptive weights to singular values, thereby removing sparse noise and redundancy while preserving the desired structural information. Furthermore, Hu et al. \cite{hu2024subspace} proposed a LLRR variant with transformed Schatten-1 penalty called TS1-LLRR, which directly adopts a nonconvex transformed function and imposes penalties on the singular values to more accurately preserve the dominant components. In addition, Qiao et al. \cite{qiao2024efficient} constructed efficient subspace clustering and feature extraction (ESCFE), which reduced the computational complexity of LLRR by using the $\ell_{2,1}$-norm and $\ell_{1,2}$-norm as convex relaxations of the nuclear norm.

Although matrix-based methods have shown promising   performance in low-rank modeling, many real-world datasets exhibit inherently high-order structures \cite{qi2017tensor}. For example, videos and color images can be represented as third-order tensors \cite{han2024nested,lin2024tensor}. Flattening such data into matrices often destroys the spatiotemporal correlations, leading to degraded performance in clustering \cite{sun2023learning}. In this regard, Fu et al. \cite{fu2016tensor} incorporated Tucker decomposition and sparse coding into LRR. However, each sample still needs to be flattened, and this may suffer from scalability limitations. Furthermore, Lu et al. \cite{lu2019tensor} proposed tensor robust principal component analysis (TRPCA) based on the tensor singular value decomposition (T-SVD), which preserves multi-dimensional structures and achieves good results in video recovery. Recently, Zhou et al. \cite{zhou2019tensor} successfully extended LRR to tensor LRR (TLRR) and provided theoretical guarantees. Case studies confirmed that TLRR has better robustness than TRPCA. Later, Du et al. \cite{du2022enhanced} regularized the $\ell_{1}$-norm and $\ell_{2,1}$-norm terms to improve robustness to complex noise, which is called enhanced TLRR (ETLRR). More recently, Ding et al. \cite{ding2025bilateral} generalized LLRR to bilateral tensor LRR (BTLRR), which can alleviate the potential performance degradation of TLRR under insufficient sampling.

\begin{figure*}[!t]
    \centering
    \includegraphics[width=17cm]{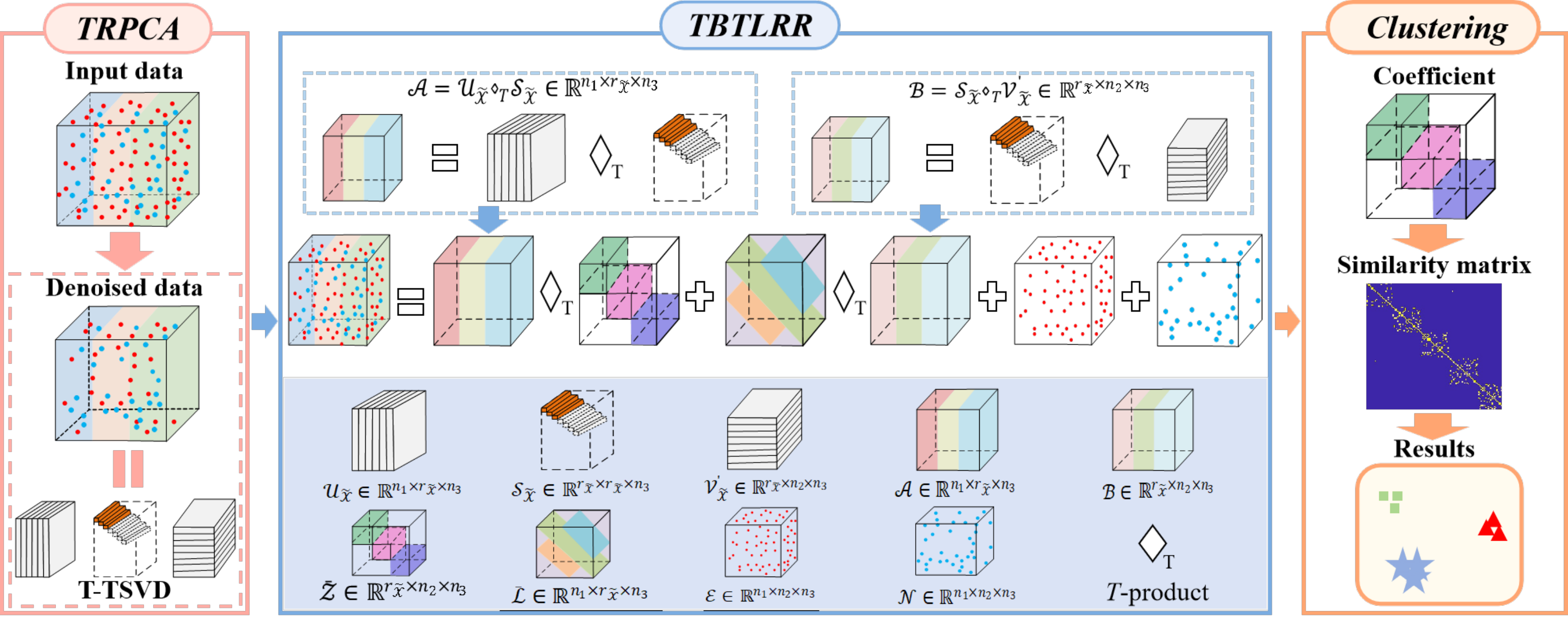}  
    \caption{Overview of the proposed TBTLRR. Firstly, TRPCA is used to remove noise from the input data. Secondly, two dictionary tensors are constructed via the T-TSVD. Thirdly, the proposed TBTLRR decomposes the data into low-rank, sparse, and Gaussian noise tensors. Finally, clustering is performed on the coefficient tensor to obtain the final results.}
    \label{fig:1}  
\end{figure*}

Note that the above TLRR and BTLRR rely on the T-SVD framework and handle the tensor nuclear norm (TNN) in the Fourier domain, which assumes that the data has an optimal representation in the frequency domain. This assumption may fail to capture global structural features when correlations are stronger in other transform domains \cite{liu2023learnable}. Furthermore, although the $\ell_{1}$-norm is widely used to characterize sparse noise, its robustness is significantly reduced when faced with mixed noise (e.g., the coexistence of sparse and Gaussian noise) \cite{chen2025adaptive}. Of course, noise can also affect the construction of the affinity matrix, causing weak diagonal dominance and lower clustering accuracy \cite{lan2023multiview}. Although LRR-based deep learning methods have achieved impressive results in many vision tasks, they require large amounts of labeled data and high computational resources \cite{xu2023deep,liu2024deep,chen2024flex}. Therefore, an important question comes to us: \textit{whether it is possible to develop a novel tensor LRR framework that can adapt to different transform domains, deal with mixed noise, and improve clustering stability.}

Motivated by these observations, we propose a data-adaptive transformed bilateral tensor LRR (TBTLRR) method to enhance the robustness in high-dimensional data clustering, as shown in Fig. \ref{fig:1}. Compared to existing work, this study offers the following four key contributions.
\begin{itemize}
    \item We construct an efficient tensor representation framework by applying a data-adaptive unitary transform along the third mode and introducing the transformed TNN, thus capturing global correlations and structural dependencies more effectively than conventional FFT-based methods.
    \item We design an enhanced denoising strategy that combines the $\ell_{1/2}$-norm for removing sparse noise and Frobenius norm for filtering out Gaussian noise, thereby improving the robustness under complex noise conditions.
    \item In algorithms, we develop an alternating direction method of multipliers (ADMM)-based scheme that incorporates low-rank subspace projections to reduce computational complexity, achieving a favorable trade-off between accuracy and efficiency.
    \item For the task of clustering, we provide a robust affinity matrix construction approach that integrates a low-rank prior with our proposed model. This improves diagonal dominance and mitigates noise interference, resulting in more reliable similarity measures.
\end{itemize}

The rest of this paper is organized as follows. Section \ref{preliminaries} reviews notations and related work. Section \ref{method} introduces our proposed model and optimization algorithm, with convergence and complexity analysis. Section \ref{Clustering} shows numerical results for clustering. Section \ref{Conclusion} concludes this paper.

\section{Preliminaries}\label{preliminaries}

This section introduces some notations and related preliminaries used throughout this paper.

\subsection{Notations and Definitions}

In this paper, tensors are represented by calligraphic uppercase letters (e.g., $\mathcal{B}$), matrices by bold uppercase letters (e.g., $\mathbf{B}$), vectors by bold lowercase letters (e.g., $\mathbf{b}$), and scalars by lowercase letters (e.g., $b$). For a third-order tensor $\mathcal{B}\in\mathbb{R}^{n_{1}\times n_{2}\times n_{3}}$, let $\mathcal{B}(i,j,k)$ denote the $(i,j,k)$-th entry, $\mathcal{B}(i,j,:)$ denote the mode-3 fiber at $(i,j)$, and $\mathcal{B}(:,:,k)$ denote the $k$-th frontal slice (also $\mathcal{B}^{(k)}$). The $\ell_1$-norm, Frobenius norm, and infinity norm are defined as
$\|\mathcal{B}\|_{1}=\sum_{i,j,k}|\mathcal{B}(i,j,k)|$,
$\|\mathcal{B}\|_{F}=(\sum_{i,j,k}|\mathcal{B}(i,j,k)|^{2})^{1/2}$,
$\|\mathcal{B}\|_{\infty}=\max_{i,j,k}|\mathcal{B}(i,j,k)|$.
In addition, the $\ell_{1/2}$-norm\footnote{Although it is not a true norm, we  still call it the norm in this paper.} is given by
$\|\mathcal{B}\|_{1/2} = ( \sum_{i,j,k} \lvert \mathcal{B}(i,j,k) \rvert^{1/2})^{2}$. Moreover, $\mathbf{I}$ is the identity matrix of an appropriate dimension. Further notations will be introduced if necessary.

\begin{Definition} [Unitary transform \cite{song2020robust}]
Let $\mathbf{T} \in \mathbb{R}^{n_3 \times n_3}$ be a unitary matrix satisfying $\mathbf{T}\mathbf{T}^{'}=\mathbf{T}^{'}\mathbf{T}=\mathbf{I}$, where $\mathbf{T}^{'}$ is the transpose of $\mathbf{T}$. The unitary transform of a third-order tensor $\mathcal{B} \in \mathbb{R}^{n_{1} \times n_{2} \times n_{3}}$ along the third dimension is composed by the product of $\mathbf{T}$ and all tubes of $\mathcal{B}$, i.e.,
\begin{equation}  
\bar{\mathcal{B}}_{T}(i, j, :) = \mathbf{T}(\mathcal{B}(i, j, :)).
\end{equation}
\end{Definition}

The original tensor $\mathcal{B}$ can be recovered from $\bar{\mathcal{B}}_{T}$ by the inverse unitary transform along the third dimension. 


\begin{Definition} [Block diagonal and fold \cite{song2020robust}]
For a third-order tensor $\bar{\mathcal{B}}_{T} \in \mathbb{R}^{n_1 \times n_2 \times n_3}$, the block diagonal form is 
\begin{equation}
\texttt{bdiag}(\bar{\mathcal{B}}_{T}) = 
\begin{bmatrix}
\bar{\mathcal{B}}_{T}^{(1)} & & & \\
& \bar{\mathcal{B}}_{T}^{(2)} & & \\
& & \ddots & \\
& & & \bar{\mathcal{B}}_{T}^{(n_3)}
\end{bmatrix}.
\end{equation}
\end{Definition}

Moreover, the tensor can be reconstructed by 
\begin{equation}
\texttt{fold}(\mathrm{bdiag}(\bar{\mathcal{B}_{T}})) = \bar{\mathcal{B}_{T}},
\end{equation}
where \texttt{fold} denotes the inverse operation of bdiag.


\begin{Definition} [$T$-product \cite{kernfeld2015tensor}]
The $T$-product between two tensors $\mathcal{B} \in \mathbb{R}^{n_1 \times n_2 \times n_3}$ and $\mathcal{C} \in \mathbb{R}^{n_2 \times n_4 \times n_3}$ is defined as 
\begin{equation}
\begin{aligned}
\mathcal{B} \diamond_T
 \mathcal{C} &= \mathbf{T}^{'}(\texttt{fold}(\texttt{bdiag}(\mathcal{\bar{B}}_{T}) \cdot \texttt{bdiag}(\mathcal{\bar{C}}_{T})))\\
 &\in \mathbb{R}^{n_1 \times n_4 \times n_3}.
\end{aligned}
\end{equation}

\end{Definition}

\begin{Definition} [Orthogonal tensor \cite{kernfeld2015tensor}]
Under a unitary transform, if the following statement holds
\begin{equation}
\mathcal{U}^{'}\diamond_T\mathcal{U}=\mathcal{U}\diamond_T\mathcal{U}^{'}=\mathcal{I},
\end{equation}
then $\mathcal{U} \in \mathbb{R}^{n \times n \times n_3}$ is called orthogonal.
\end{Definition}

\begin{Definition} [$f$-diagonal \cite{kolda2009tensor}]
A tensor is called $f$-diagonal if every frontal slice of the tensor is a diagonal matrix.
\end{Definition}

We now introduce TNNs, Transformed T-SVD (T-TSVD), and Transformed TNNs (TTNNs).

\begin{Definition} [TNN \cite{kolda2009tensor}]
Given a third-order tensor $\mathcal{B} \in \mathbb{R}^{n_1 \times n_2 \times n_3}$, its  tensor nuclear norm (TNN) is defined as
\begin{equation}
\| \mathcal{B} \|_{\text{TNN}} = \frac{1}{n_3} \sum_{k=1}^{n_3} \| \bar{\mathcal{B}}^{(k)} \|_{*},
\end{equation}
where $\| \bar{\mathcal{B}}^{(k)} \|_{*}$ is the nuclear norm of $\bar{\mathcal{B}}^{(k)}$ and $\bar{\mathcal{B}}$ is obtained by the FFT of $\mathcal{B}$.
\end{Definition}

\begin{Definition} [T-TSVD \cite{kernfeld2015tensor}]
Given a third-order tensor $\mathcal{B} \in \mathbb{R}^{n_1 \times n_2 \times n_3}$, it admits the transformed tensor singular value decomposition (T-TSVD) in the form of
\begin{equation}
\mathcal{B} = \mathcal{U} \diamond_{T} \mathcal{S} \diamond_{T} \mathcal{V}^{'},
\end{equation}
where $\mathcal{U} \in \mathbb{R}^{n_1 \times r \times n_3}$ and $\mathcal{V} \in \mathbb{R}^{n_2 \times r \times n_3}$ are the orthogonal tensors, and $\mathcal{S} \in \mathbb{R}^{r \times r \times n_3}$ is an $f$-diagonal tensor.
\end{Definition}

\begin{Definition} [TTNN \cite{kernfeld2015tensor}]
Given a third-order tensor $\mathcal{B} \in \mathbb{R}^{n_1 \times n_2 \times n_3}$, its transformed TNN (TTNN) is defined as
\begin{equation}
\| \mathcal{B} \|_{\otimes} = \frac{1}{n_3} \sum_{k=1}^{n_3} \| \bar{\mathcal{B}}_{T}^{(k)} \|_{*},
\end{equation}
where $\| \bar{\mathcal{B}}_{T}^{(k)} \|_{*}$ is the nuclear norm of $\bar{\mathcal{B}}_{T}^{(k)}$.
\end{Definition}

\subsection{Related Work}

Given the observed tensor data $\mathcal{X} \in \mathbb{R}^{n_1 \times  n_2 \times n_3}$, TLRR \cite{zhou2019tensor} aims to find a low-rank tensor $\mathcal{Z} \in \mathbb{R}^{n_1 \times n_2 \times n_3}$ and a sparse tensor $\mathcal{E} \in \mathbb{R}^{n_1 \times n_2 \times n_3}$ such that 
\begin{equation}\label{TLRR}
\begin{aligned}
\min_{\mathcal{Z}, \mathcal{E}}\quad&\| \mathcal{Z} \|_{\text{TNN}} + \lambda \| \mathcal{E} \|_1 \\
\textrm{s.t.} \quad&\mathcal{X} = \mathcal{A} * \mathcal{Z} + \mathcal{E},
\end{aligned}
\end{equation}
where $\lambda>0$ is the regularization parameter to balance the low rank and sparsity. $\mathcal{A}\in \mathbb{R}^{n_1 \times n_2 \times n_3}$ is the learned dictionary and a common choice is $\mathcal{X}$, which is known as self-representation. Although TLRR performs well on data that is complete and has little noise, its performance may degrade in the presence of missing or noisy data.

To address this limitation, BTLRR \cite{ding2025bilateral} is proposed by incorporating latent low-rank terms to capture hidden structural dependencies. The optimization model is
\begin{equation}\label{BTLRR}
\begin{aligned}
\min_{\mathcal{Z}, \mathcal{L}, \mathcal{E}} \quad &\| \mathcal{Z} \|_{\text{TNN}} + \| \mathcal{L} \|_{\text{TNN}} + \lambda \| \mathcal{E} \|_1 \\
\text{s.t.}\quad~ &\mathcal{X} = \tilde{\mathcal{X}} * \mathcal{Z} + \mathcal{L} * \tilde{\mathcal{X}} + \mathcal{E},
\end{aligned}
\end{equation}
where $\tilde{\mathcal{X}}\in \mathbb{R}^{n_1 \times n_2 \times n_3}$ is the dictionary obtained from $\mathcal{X}$ via denoising, which serves as a cleaner basis for representing the data. Compared with \eqref{TLRR}, the latent low-rank tensor $\mathcal{Z}$ is designed to capture the similarities between samples, while  $\mathcal{L}$ can be interpreted as modeling the relationships between features. It is validated that these two structural representations complement each other, achieving satisfactory clustering and recovery performance.

\section{Methodology} \label{method}

This section first introduces our new model, and then develops an efficient optimization algorithm, along with rigorous convergence analysis and complexity analysis.

\begin{figure}[t]
    \centering
    \subfigure[]{
        \label{a1}
        \includegraphics[height=2.5cm]{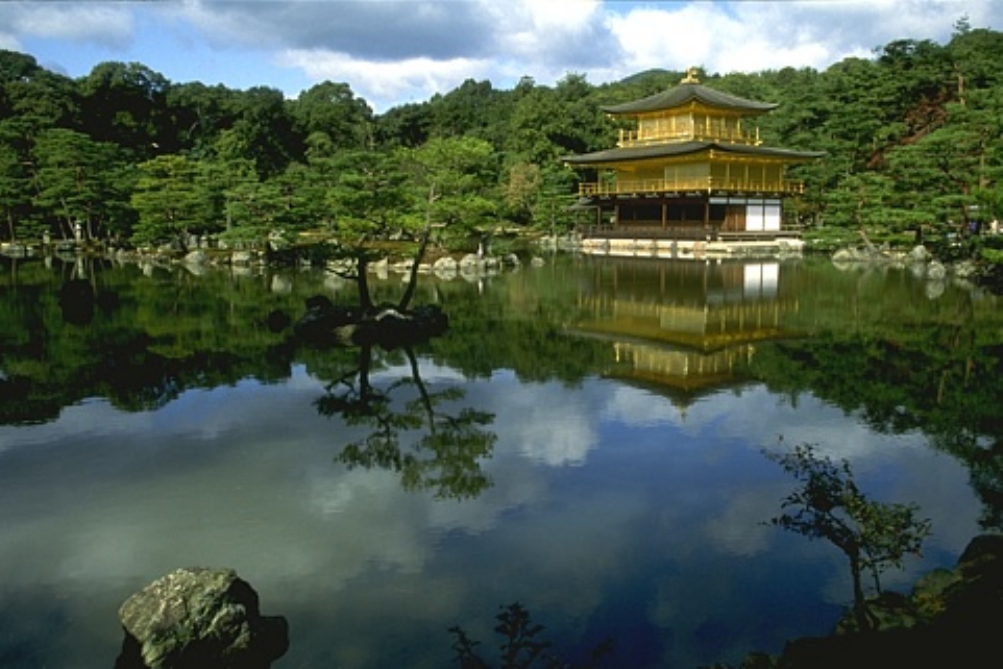}
    } \hspace{2mm}
    \subfigure[]{
        \label{b1}
        \includegraphics[height=2.5cm]{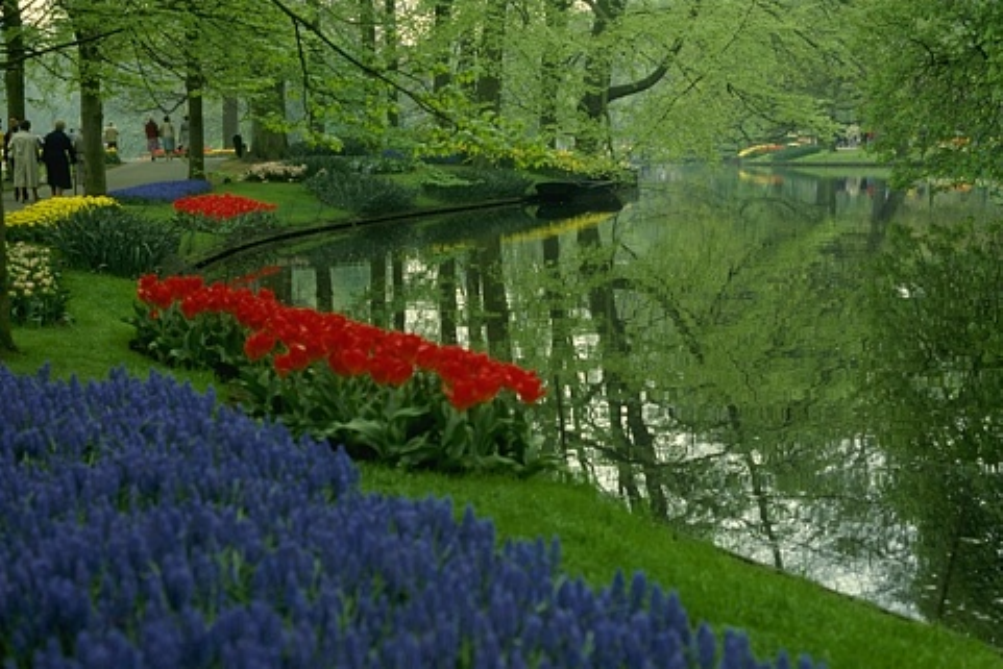}
    }
    \subfigure[]{
        \label{c1}
        \includegraphics[height=3cm]{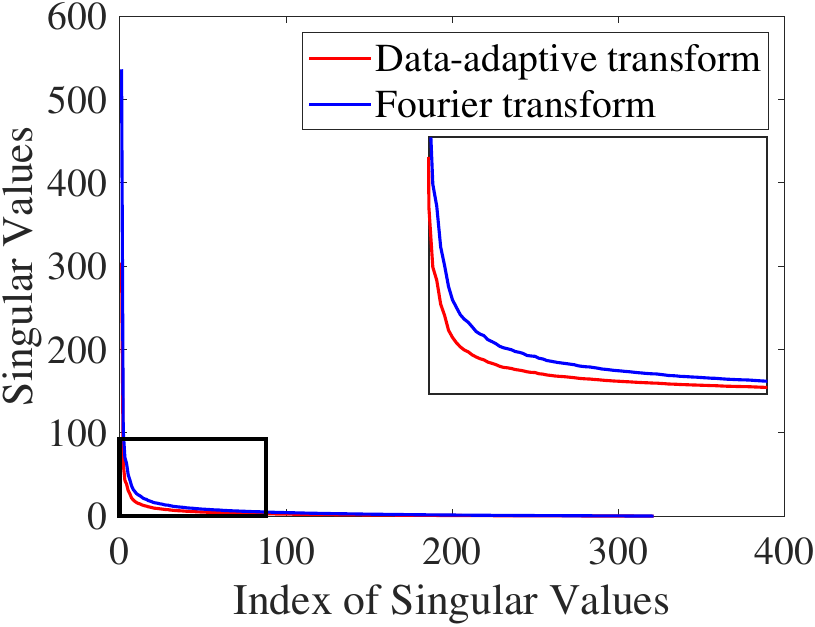}
    } \hspace{-1mm}
    \subfigure[]{
        \label{d1}
        \includegraphics[height=3cm]{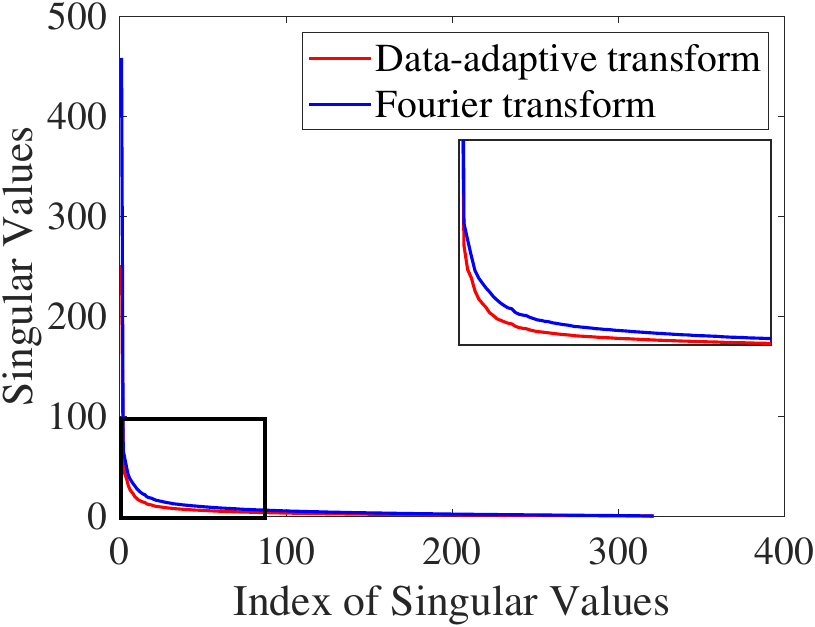}
    }
    \caption{Illustration of the low tubal-rank property, where (a)-(b) are the selected images from the Berkeley Segmentation dataset, (c)-(d) are the corresponding singular value sum distributions.}
    \label{fig:tubal-rank}
\end{figure}

\subsection{New Model}
Recall that TLRR and BTLRR only consider the nuclear norm defined in the fast Fourier transform (FFT) domain. As illustrated in Fig. \ref{fig:tubal-rank}, singular values in non-FFT transform domains decay significantly faster than those in the FFT domain, indicating a stronger low-rank property in the transformed representation. In addition, real-world data are often corrupted by both sparse outliers and dense Gaussian noise, which may degrade performance in the subsequent tasks.

Therefore, we propose the transformed bilateral tensor low-rank representation (TBTLRR), which is formulated as 
\begin{equation}
\begin{aligned}
\label{eq:ttnn_model}
\min_{\mathcal{Z}, \mathcal{L}, \mathcal{E}, \mathcal{N}}
\quad &\|\mathcal{Z}\|_{\otimes} + \|\mathcal{L}\|_{\otimes}
+ \lambda\|\mathcal{E}\|_{{1/2}} + \beta\|\mathcal{N}\|_F^2 \\
\text{s.t.}~~\quad
&\mathcal{X} = \tilde{\mathcal{X}} \diamond_{T} \mathcal{Z}
+ \mathcal{L} \diamond_{T} \tilde{\mathcal{X}}
+ \mathcal{E} + \mathcal{N},
\end{aligned}
\end{equation}
where $\|\cdot\|_{\otimes}$ denotes the TTNN, $\|\cdot\|_{{1/2}}$ denotes the $\ell_{1/2}$-norm, and $\lambda, \beta>0$ are the trade-off parameters.

Compared with \eqref{BTLRR}, the proposed TBTLRR extends the bilateral low-rank modeling paradigm of BTLRR from the fixed FFT domain to arbitrary unitary transform domains. This flexibility allows the transform to be adaptively selected or learned according to data characteristics. Besides, the $\ell_{1/2}$-norm regularization is integrated to filter out sparse outliers, which is more effective than $\ell_{1}$-norm, as stated in  \cite{liu2024efficient}. Moreover, the Frobenius norm regularization is added to eliminate Gaussian noise. As a result, our model can effectively handle heterogeneous noise and is more robust.

\begin{Remark}
In fact, the choice of the unitary transformation plays a vital role in tensor representation. 
An appropriate transformation enables the TTNN to more effectively capture the underlying low-rank structures. 
In this paper, we employ a data-adaptive transform in the clustering task.
In particular, the observed tensor $\mathcal{X}$ is unfolded along the third mode to form $\mathbf{L}$, followed by the SVD, i.e., $\mathbf{L}=\mathbf{U}\mathbf{S}\mathbf{V}^{'}$. The unitary matrix $\mathbf{T}=\mathbf{U}^{'}$ is used as the transform in TTNN for clustering, better capturing the low-rank structure.
\end{Remark}

\subsection{Optimization Algorithm}
This section develops an efficient algorithm for solving \eqref{eq:ttnn_model}. 
Inspired by \cite{ding2025bilateral}, the subspace projection technique is adopted to improve the scalability on large-scale data. 
Let the skinny T-TSVD of $\tilde{\mathcal{X}}$ be 
$\tilde{\mathcal{X}}=\mathcal{U}_{\tilde{\mathcal{X}}}\diamond_{T}\mathcal{S}_{\tilde{\mathcal{X}}}\diamond_{T}\mathcal{V}_{\tilde{\mathcal{X}}}^{'}$ 
and set $r_{\tilde{\mathcal{X}}}=\operatorname{rank}(\tilde{\mathcal{X}})$. 

Define 
\begin{equation}
\begin{aligned}
\mathcal{A}&= \mathcal{U}_{\tilde{\mathcal{X}}}\diamond_{T}\mathcal{S}_{\tilde{\mathcal{X}}}\in\mathbb{R}^{n_{1}\times r_{\tilde{\mathcal{X}}}\times n_{3}},\\
\mathcal{B}&= \mathcal{S}_{\tilde{\mathcal{X}}}\diamond_{T}\mathcal{V}_{\tilde{\mathcal{X}}}^{'}\in\mathbb{R}^{r_{\tilde{\mathcal{X}}}\times n_{2}\times n_{3}}.
\end{aligned}
\end{equation}

Introduce the projected variables
\begin{equation}
\begin{aligned}
\bar{\mathcal{Z}}&= \mathcal{V}_{\tilde{\mathcal{X}}}^{'}\diamond_{T}\mathcal{Z}\in\mathbb{R}^{r_{\tilde{\mathcal{X}}}\times n_{2}\times n_{3}},\\
\bar{\mathcal{L}}&= \mathcal{L}\diamond_{T}\mathcal{U}_{\tilde{\mathcal{X}}}\in\mathbb{R}^{n_{1}\times r_{\tilde{\mathcal{X}}}\times n_{3}}.
\end{aligned}
\end{equation}

Then, \eqref{eq:ttnn_model} is equivalently rewritten as
\begin{equation}
\begin{aligned}
\label{eq:8}
 \min_{\bar{\mathcal{Z}}, \bar{\mathcal{L}}, \mathcal{N},\mathcal{E}} \quad &\|\bar{\mathcal{Z}}\|_{\otimes} + \|\bar{\mathcal{L}}\|_{\otimes}+\lambda\|\mathcal{E}\|_{1/2}+\beta \|\mathcal{N}\|_{F}^2 \\
\textrm{s.t.} ~~~\quad &\mathcal{X} = \mathcal{A} \diamond_{T} \bar{\mathcal{Z}} + \bar{\mathcal{L}} \diamond_{T} \mathcal{B} + \mathcal{E}+\mathcal{N}.
\end{aligned}
\end{equation}

According to the alternating direction method
of multipliers (ADMM), two variables $\mathcal{J}\in \mathbb{R}^{r_{\tilde{\mathcal{X}}} \times n_2 \times n_3}$, $\mathcal{T}\in \mathbb{R}^{n_1 \times r_{\tilde{\mathcal{X}}} \times n_3}$ are introduced, which derives
\begin{equation}
\begin{aligned}
\label{eq:14}
\min_{\mathcal{J}, \bar{\mathcal{Z}},  \mathcal{T}, \bar{\mathcal{L}},\mathcal{N},\mathcal{E}} \quad & \|\bar{\mathcal{Z}}\|_{\otimes} + \|\bar{\mathcal{L}}\|_{\otimes}+\lambda\|\mathcal{E}\|_{1/2}+\beta\|\mathcal{N}\|_{F}^2 \\
\textrm{s.t.} ~~\quad\quad &\mathcal{X} = \mathcal{A} \diamond_{T} \bar{\mathcal{Z}} + \bar{\mathcal{L}} \diamond_{T} \mathcal{B} + \mathcal{E}+\mathcal{N},\\
&\bar{\mathcal{Z}}=\mathcal{J}, \bar{\mathcal{L}}=\mathcal{T},
\end{aligned}
\end{equation}
and its augmented Lagrangian function is 
\begin{equation}
\begin{aligned}
\label{eq:L}
    & \mathcal{L}_{\mu}({\mathcal{J}, \bar{\mathcal{Z}},  \mathcal{T}, \bar{\mathcal{L}},\mathcal{N},\mathcal{E}, \mathcal{P}, \mathcal{G}, \mathcal{W}}) \\
&=\|\mathcal{J}\|_{\otimes}+\|\mathcal{T}\|_{\otimes}+\lambda \|\mathcal{E}\|_{1/2}+\beta\|\mathcal{N}\|_{F}^2 \\
&+\langle\mathcal{P},\mathcal{X}-\mathcal{A}\diamond_{T}\bar{\mathcal{Z}}-\bar{\mathcal{L}}\diamond_{T}\mathcal{B}-\mathcal{E}-\mathcal{N}\rangle\\
&+\frac{\mu}{2} \|\mathcal{X} - \mathcal{A}\diamond_{T} \bar{\mathcal{Z}} - \bar{\mathcal{L}}\diamond_{T} \mathcal{B} - \mathcal{E}-\mathcal{N}\|_F^2\\
&+\langle\mathcal{G},\bar{\mathcal{Z}}-\mathcal{J}\rangle+\frac{\mu}{2} \|\bar{\mathcal{Z}} - \mathcal{J}\|_F^2\\
&+\langle\mathcal{W},\bar{\mathcal{L}}-\mathcal{T}\rangle+\frac{\mu}{2} \|\bar{\mathcal{L}} - \mathcal{T}\|_F^2.
\end{aligned}
\end{equation}
Here, $\mathcal{P}\in \mathbb{R}^{n_1 \times n_2 \times n_3}, \mathcal{G}\in \mathbb{R}^{r_{\tilde{\mathcal{X}}} \times n_2 \times n_3}, \mathcal{W}\in \mathbb{R}^{n_1 \times r_{\tilde{\mathcal{X}}} \times n_3}$ are the Lagrange multipliers and $\mu>0$ is the penalty parameter. Therefore, the objective function can be minimized by alternately solving one variable while fixing the other variables.

\subsubsection{Update $\mathcal{J}$ while fix others}
\begin{equation}
\label{eq:J_subproblem}
\mathcal{J}_{t+1}
=\arg\min_{\mathcal{J}}~ \|\mathcal{J}\|_{\otimes}
+\frac{\mu}{2}\|\mathcal{J}-(\bar{\mathcal{Z}}_{t}+\frac{\mathcal{G}_{t}}{\mu})\|_{F}^{2}.
\end{equation}


Let the T-TSVD of $\bar{\mathcal{Z}}_{t}+\frac{\mathcal{G}_{t}}{\mu}$ be $\mathcal{U}_{t}\diamond_{T}\,\mathcal{S}_{t}\,\diamond_{T}\mathcal{V}_{t}^{'}$.
Then the closed-form solution is
\begin{equation}
\label{eq:J_update}
\mathcal{J}_{t+1}
=\mathcal{U}_{t}\diamond_{T}\,\mathcal{D}_{1/\mu}(\mathcal{S}_{t})\,\diamond_{T}\mathcal{V}_{t}^{'},
\end{equation}
where $\mathcal{D}_{1/\mu}(\cdot)$ denotes the transformed soft-thresholding on the singular tubes, i.e.,
\begin{equation}
\label{eq:D_operator}
\mathcal{D}_{1/\mu}(\mathcal{S})
= \mathbf{T}^{'}[\max(\bar{\mathcal{S}}_{T}-1/\mu,\,0)].
\end{equation}

\subsubsection{Update $\bar{\mathcal{Z}}$ while fix others}
\begin{equation}
\label{eq:19}
\begin{aligned}
\bar{\mathcal{Z}}_{t+1}
&= \arg\min_{\bar{\mathcal{Z}}}~
   \frac{\mu}{2}\,\|\bar{\mathcal{Z}} - (\mathcal{J}_{t+1}-\frac{\mathcal{G}_{t}}{\mu})\|_F^2 \\
 &  + \frac{\mu}{2}\|
      \mathcal{A}\diamond_{T} \bar{\mathcal{Z}}
      - (\mathcal{C}_1+\frac{\mathcal{P}_{t}}{\mu})\|_{F}^{2},
\end{aligned}
\end{equation}
where $\mathcal{C}_1=\mathcal{X}- \bar{\mathcal{L}}_{t} \diamond_{T} \mathcal{B} - \mathcal{E}_{t} - \mathcal{N}_{t}$. It is easy to see  that
\begin{equation}
\begin{aligned}
\label{eq:20}
\bar{\mathcal{Z}}_{t+1} &= (\mathcal{I}+\mathcal{A}^{'}\diamond_{T}\mathcal{A})^{-1}\diamond_{T}( \mathcal{A}^{'} \diamond_{T}\mathcal{C}_1  \\
&+\mathcal{J}_{t+1}+\frac{\mathcal{A}^{'}\diamond_{T} \mathcal{P}_{t} - \mathcal{G}_{t}}{\mu}).
\end{aligned}
\end{equation}

\subsubsection{Update $\mathcal{T}$ while fix others}
\begin{equation}
\label{eq:T_subproblem}
\mathcal{T}_{t+1}
=\arg\min_{\mathcal{T}}~ \|\mathcal{T}\|_{\otimes}
+\frac{\mu}{2}\|\mathcal{T}-(\bar{\mathcal{L}}_{t}+\frac{\mathcal{W}_{t}}{\mu})\|_{F}^{2}.
\end{equation}

Similarly to updating $\mathcal{J}$, let
\begin{equation}
\label{eq:T_ttsvd}
\bar{\mathcal{L}}_{t}+\frac{\mathcal{W}_{t}}{\mu}
=\tilde{\mathcal{U}}_{t}\diamond_{T}\,\tilde{\mathcal{S}}_{t}\,\diamond_{T}\tilde{\mathcal{V}}_{t}^{'},
\end{equation}
and thus the solution is
\begin{equation}
\label{eq:T_update_closed}
\mathcal{T}_{t+1}
=\tilde{\mathcal{U}}_{t}\diamond_{T}\,\mathcal{D}_{1/\mu}(\tilde{\mathcal{S}}_{t})\,\diamond_{T}\tilde{\mathcal{V}}_{t}^{'}.
\end{equation}

\subsubsection{Update $\bar{\mathcal{L}}$ while fix others}
\begin{equation}
\label{eq:21}
\begin{aligned}
\bar{\mathcal{L}}_{t+1}
&= \arg \min_{\bar{\mathcal{L}}}~\ 
   \frac{\mu}{2}\,
   \|\bar{\mathcal{L}} - (\mathcal{T}_{t+1}-\frac{\mathcal{W}_{t}}{\mu})\|_F^2\\
&+\frac{\mu}{2}\|\bar{\mathcal{L}}\diamond_{T}\mathcal{B}-(\mathcal{C}_2+\frac{\mathcal{P}_{t}}{\mu})\|_F^2, \\
\end{aligned}
\end{equation}
where $\mathcal{C}_2=\mathcal{X} - \mathcal{A}\diamond_{T}\bar{\mathcal{Z}}_{t+1} - \mathcal{E}_{t}-\mathcal{N}_{t}$. Setting its derivative to zero, it obtains
\begin{equation}
\begin{aligned}
\label{eq:22}
\bar{\mathcal{L}}_{t+1} &= ( \mathcal{C}_2\diamond_{T}\mathcal{B}^{'}+ \mathcal{T}_{t+1} 
+ \frac{\mathcal{P}_{t} \diamond_{T}\mathcal{B}^{'} - \mathcal{W}_{t}}{\mu})\\
&\diamond_{T}(\mathcal{I} + \mathcal{B}\diamond_{T} \mathcal{B}^{'})^{-1}.
\end{aligned}
\end{equation}

\subsubsection{Update $\mathcal{N}$ while fix others}
\begin{equation}
\label{eq:23}
\begin{aligned}
\mathcal{N}_{t+1}
= \arg\min_{\mathcal{N}}~ 
   \beta\|\mathcal{N} \|_F^2
 + \frac{\mu}{2}\|\mathcal{N} - (\mathcal{C}_{3}+\frac{\mathcal{P}_{t}}{\mu})\|_F^2 ,
\end{aligned}
\end{equation}
where  $ \mathcal{C}_3=\mathcal{X}- \mathcal{A} \diamond_{T} \bar{\mathcal{Z}}_{t+1}-\bar{\mathcal{L}}_{t+1} \diamond_{T} \mathcal{B}- \mathcal{E}_{t}$. The solution is
\begin{equation}
\begin{aligned}
\label{eq:24}
\mathcal{N}_{t+1} = \frac{\mathcal{P}_{t}+{\mu}\mathcal{C}_3}{2\beta+\mu}.
\end{aligned}
\end{equation}

\subsubsection{Update $\mathcal{E}$ while fix others}
\begin{equation}
\label{eq:E_subproblem}
\mathcal{E}_{t+1}
=\arg\min_{\mathcal{E}}~\lambda\|\mathcal{E}\|_{1/2}
+\frac{\mu}{2}\|\mathcal{E}-(\mathcal{C}_{4}+\frac{\mathcal{P}_{t}}{\mu})\|_{F}^{2},
\end{equation}
where $\mathcal{C}_{4}= \mathcal{X}-\mathcal{A}\diamond_{T}\bar{\mathcal{Z}}_{t+1}
-\bar{\mathcal{L}}_{t+1}\diamond_{T}\mathcal{B}-\mathcal{N}_{t+1}$. Let $\mathcal{Y}_{t}=\mathcal{C}_{4}+\frac{\mathcal{P}_{t}}{\mu}$ and $\alpha=\lambda/\mu$.
Since \eqref{eq:E_subproblem} is separable element-wisely, the closed-form solution is
\begin{equation}
\label{eq:E_update}
\mathcal{E}_{t+1}=\mathcal{H}_{\alpha}(\mathcal{Y}_{t}),
\end{equation}
where the half-thresholding function $\mathcal{H}_{\alpha}(\cdot)$ is
\begin{equation}
\label{eq:half_shrink}
\mathcal{H}_{\alpha}(y)=
\begin{cases}
0, & |y|\le \tau(\alpha),\\[3pt]
\operatorname{sgn}(y)\,\dfrac{2}{3}|y|[1+\cos\varphi(y,\alpha)], & |y|>\tau(\alpha),
\end{cases}
\end{equation}
with
\begin{equation}
\label{eq:tau_phi}
\tau(\alpha)=\tfrac{3}{2}\alpha^{2/3} \quad \textrm{and} \quad
\varphi(y,\alpha)=\tfrac{2}{3}\arccos~(\smash{\dfrac{3\sqrt{3}\,\alpha}{4|y|^{3/2}}}).
\end{equation}

\subsubsection{Update multipliers $\mathcal{P}_{t+1}, 
\mathcal{G}_{t+1}, \mathcal{W}_{t+1}$ by}
\begin{equation}\label{eq:mul}
    \begin{cases}
        \mathcal{P}_{t+1} = \mathcal{P}_{t} + \mu ( \mathcal{X} - \mathcal{A} \diamond_T \bar{\mathcal{Z}}_{t+1} - \bar{\mathcal{L}}_{t+1} \diamond_T \mathcal{B} - \mathcal{E}_{t+1} - \mathcal{N}_{t+1}), \\
        \mathcal{G}_{t+1} = \mathcal{G}_{t} + \mu ( \bar{\mathcal{Z}}_{t+1} - \mathcal{J}_{t+1} ), \\
        \mathcal{W}_{t+1} = \mathcal{W}_{t} + \mu ( \bar{\mathcal{L}}_{t+1} - \mathcal{T}_{t+1} ),
    \end{cases}
\end{equation}
and $\mu=\min(\rho\mu,\mu_{max})$.
The detailed iterative scheme is summarized in Algorithm \ref{algorithm：1}, and the convergence conditions are
        \begin{equation}\label{conv}
        \begin{cases}
            \| \bar{\mathcal{Z}}_{t} - \mathcal{J}_{t}\|_{\infty}<\varepsilon,\\
            \|\bar{\mathcal{L}}_{t} - \mathcal{T}_{t}\|_{\infty}< \varepsilon,\\
            \|\mathcal{X} - \mathcal{A} \diamond_{T} \bar{\mathcal{Z}}_{t} - \bar{\mathcal{L}}_{t} \diamond_{T} \mathcal{B} - \mathcal{E}_{t} - \mathcal{N}_{t}\|_{\infty}< \varepsilon.
        \end{cases}
   \end{equation}

\begin{algorithm}[t]
\caption{ADMM for solving  \eqref{eq:14}}
\label{algorithm：1}
 \textbf{Input:} $\mathcal{X}$, $\lambda$, $\beta$, $\mu=10^{-7}$, $\mu_{max}=10^{7}$, $\rho=1.5$, $\varepsilon=10^{-7}$\\
 \textbf{Initialize:} $\bar{\mathcal{Z}}=\mathcal{J}=0$, $\bar{\mathcal{L}}=\mathcal{T}=0$, $\mathcal{E}=\mathcal{N}=0$, $\mathcal{P}=0$, $\mathcal{G}=0$, $\mathcal{W}=0$\\
 \textbf{While} not converged \textbf{do}	
 \begin{algorithmic}[1]
        \STATE Update $\mathcal{J}_{t+1}$ via \eqref{eq:J_update}
        \STATE Update $\bar{\mathcal{Z}}_{t+1}$ via \eqref{eq:20} 
        \STATE Update $\mathcal{T}_{t+1}$ via \eqref{eq:T_update_closed}
        \STATE Update $\bar{\mathcal{L}}_{t+1}$ via \eqref{eq:22}
        \STATE Update $\mathcal{N}_{t+1}$ via \eqref{eq:23}
        \STATE Update $\mathcal{E}_{t+1}$ via \eqref{eq:E_subproblem}
        \STATE Update $\mathcal{P}_{t+1}, \mathcal{G}_{t+1}, \mathcal{W}_{t+1}$ by \eqref{eq:mul} 
        \STATE Check convergence by \eqref{conv}
\end{algorithmic}
\textbf{End While}\\
 \textbf{Output:} $\mathcal{Z}=\mathcal{V}_{\mathcal{\tilde{\mathcal{X}}}}\diamond_{T}\bar{\mathcal{Z}}$
\end{algorithm}

\subsection{Convergence Analysis}
For convenience, denote
\begin{equation}
\{\mathcal{H}_t\}=\{\mathcal{J}, \bar{\mathcal{Z}},  \mathcal{T}, \bar{\mathcal{L}}, \mathcal{N}, \mathcal{E}, \mathcal{P}, \mathcal{G}, \mathcal{W}\}. 
\end{equation}
Following a similar line as \cite{yang2022robust}, the bounded property is easy to derive, so the proof is omitted.
\begin{Lemma}\label{bouned}
Assume that $\{\mathcal{H}_t\}$ is a sequence generated by Algorithm \ref{algorithm：1}. Then $\{\mathcal{H}_{t}\}$ is bounded.
\end{Lemma}

From the Bolzano--Weierstrass theorem \cite{bartle2000introduction}, the generated sequence $\{\mathcal{H}_t\}$ admits at least one limit point, say
$\mathcal{H}_{*}$.
Because the primal residuals vanish along the sequence, any limit point is primal-feasible.
Passing to the limit in the first-order optimality conditions of the subproblems yields that
$\mathcal{H}_{*}$ satisfies the Karush-Kuhn-Tucker (KKT) conditions for  \eqref{eq:14}.
Thus, any accumulation point of $\{\mathcal{H}_t\}$ is a KKT  point of  \eqref{eq:14}.

With the help of the Kurdyka–Lojasiewicz (KL) property \cite{attouch2013convergence}, we establish the convergence theorem of Algorithm \ref{algorithm：1}. 
\begin{Theorem}\label{Theorem:3.1}
Assume that $\{\mathcal{H}_t\}$ is a sequence generated by Algorithm \ref{algorithm：1}. Then $\{ \mathcal{H}_{t} \}$ is a Cauchy sequence and converges to a critical point of \eqref{eq:14}.
\end{Theorem}
\begin{IEEEproof}
Let the residuals be $ r^{(1)}_{t} = \bar{\mathcal{Z}}_{t} - \mathcal{J}_{t}$, $r^{(2)}_{t} = \bar{\mathcal{L}}_{t} - \mathcal{T}_{t}$, $r^{(3)}_{t} = \mathcal{X} - \mathcal{A} \diamond_{T} \bar{\mathcal{Z}}_{t} - \bar{\mathcal{L}}_{t} \diamond_{T} \mathcal{B} - \mathcal{E}_{t} - \mathcal{N}_{t}.$
Define 
\begin{equation}\label{eq:deltaH}
\begin{aligned}
\Delta \mathcal{H}_{t+1} = 
(
    \mathcal{J}_{t+1} - \mathcal{J}_t, \, 
    \bar{\mathcal{Z}}_{t+1} - \bar{\mathcal{Z}}_t, \, 
    \mathcal{T}_{t+1} - \mathcal{T}_t, \, \\
    \bar{\mathcal{L}}_{t+1} - \bar{\mathcal{L}}_t, \, 
    \mathcal{N}_{t+1} - \mathcal{N}_t, \, 
    \mathcal{E}_{t+1} - \mathcal{E}_t)
\end{aligned}
\end{equation}
and 
\begin{equation}\label{eq:Psi-def}
\begin{aligned}
\Phi_t &= \mathcal L_{\mu_t}(\mathcal{J}_{t},\bar{\mathcal {Z}}_{t},\mathcal{T}_{t},\bar{\mathcal{L}}_{t},\mathcal{N}_{t},\mathcal{E}_{t},\mathcal{P}_{t},\mathcal{G}_{t},\mathcal{W}_{t})\\
&+\frac{1}{2\mu_t}(\|\mathcal{P}_{t}\|_F^2+\|\mathcal {G}_{t}\|_F^2+\|\mathcal{W}_{t}\|_F^2).
\end{aligned}
\end{equation}

Based on the first-order optimality conditions and the sufficient decrease property of each proximal subproblem induced by the quadratic penalties, there exist constants $c_1, c_2 > 0$ and a  sequence $\{\varepsilon_t\}$ such that
\begin{equation}\label{eq:suff-dec}
\begin{aligned}
\Phi_t-\Phi_{t+1}&\ge 
c_1\|\Delta\mathcal H_{t+1}\|_F^2\\
&+c_2\,\mu_t(\|r^{(1)}_{t+1}\|_F^2+\|r^{(2)}_{t+1}\|_F^2+\|r^{(3)}_{t+1}\|_F^2)
-\varepsilon_t.
\end{aligned}
\end{equation}
This, together with the boundedness of the multipliers, derives 
$\sum_{t=0}^{\infty}\varepsilon_t < \infty.$ 
Summing \eqref{eq:suff-dec} from \(t = 0\) and noting that \(\{\Phi_t\}\) is bounded, it is to see that
\begin{equation}\label{eq:summable-l2}
\begin{aligned}
&\sum_{t=0}^{\infty}\|\Delta\mathcal H_{t+1}\|_F^2 < \infty,\\
&\sum_{t=0}^{\infty}\mu_t(\|r^{(1)}_{t+1}\|_F^2+\|r^{(2)}_{t+1}\|_F^2+\|r^{(3)}_{t+1}\|_F^2) < \infty.
\end{aligned}
\end{equation}
Therefore, there exists \(L>0\) such that
\begin{equation}\label{eq:rel-err}
\mathrm{dist}(0,\partial \Phi_{t+1})\le L\|\Delta\mathcal H_{t+1}\|_F.
\end{equation}
Since \(\|\cdot\|_{\otimes}\), \(\|\cdot\|_{{1/2}}\), and the Frobenius norm are semi-algebraic functions, \(\Phi\) satisfies the Kurdyka–Lojasiewicz (KL) property. As a result, there exist \(\theta \in [0,1)\) and a concave function \(E(s) = \tfrac{1}{1-\theta}s^{1-\theta}\) such that
\begin{equation}\label{eq:KL}
E'(\Phi_t-\Phi_*)\ \mathrm{dist}(0, \partial \Phi_t)\ge 1.
\end{equation}
By substituting \eqref{eq:rel-err} into \eqref{eq:KL} and combining it with \eqref{eq:suff-dec}, the iteration has finite length, i.e.,
\begin{equation}\label{eq:finite-length}
\sum_{t=0}^{\infty}\|\Delta\mathcal H_{t+1}\|_F < \infty.
\end{equation}

Therefore, for any \(\varepsilon > 0\), there exists \(T\) such that for all \(s > t \ge T\), it holds
\begin{equation}\label{eq:cauchy}
\|\mathcal H_s-\mathcal H_t\|_F\le\sum_{k=t}^{s-1}\|\Delta\mathcal H_{k+1}\|_F < \varepsilon,
\end{equation}
which shows that the sequence \(\{\mathcal H_ {t} \}\) is a Cauchy sequence. The proof is completed. 
\end{IEEEproof}

\begin{table}[t]
\centering
\caption{Detailed information of the selected datasets.}
\label{tab:data}
\renewcommand{\arraystretch}{1.3}
\begin{tabular}{|c|c|c|c|c|}
\hline
Category & Datasets & Samples & Classes & Sizes \\
\hline
\hline
\multirow{4}{*}{Small} 
& ORL             & 400   & 40 & $32 \times 32$ \\
\cline{2-5}
& Umist           & 480   & 20 & $32 \times 32$ \\
\cline{2-5}
& Extended YaleB  & 840   & 28 & $80 \times 60$ \\
\cline{2-5}
& UCSD            & 896   & 10 & $36 \times 54$ \\
\cline{2-5}
\hline
\multirow{2}{*}{Medium} 
& Notting-Hill    & 1494  & 5  & $60 \times 60$ \\
\cline{2-5}
& MSTAR           & 2425  & 10 & $32 \times 32$ \\
\hline
\multirow{2}{*}{Large} 
& GTSRB           & 6000  & 12 & $32\times 32$ \\
\cline{2-5}
& MNIST           & 10000 & 10  & $28 \times 28$ \\
\hline
\end{tabular}
\end{table}

\subsection{Complexity Analysis}
The computational complexity of each step in Algorithm \ref{algorithm：1} is analyzed below. Let's assume that $r_{\tilde{\mathcal{X}}}\ll \min(n_1,n_2)$. Updating $\mathcal{J}_{t+1}$ requires $\mathcal{O}(r_{\tilde{\mathcal{X}}}\,n_2 n_3(n_2+n_3))$. The update for $\mathcal{T}_{t+1}$ costs $\mathcal{O}(r_{\tilde{\mathcal{X}}}\,n_1 n_3(r_{\tilde{\mathcal{X}}}+n_3))$. For $\bar{\mathcal{Z}}_{t+1}$, the complexity is $\mathcal{O}(r_{\tilde{\mathcal{X}}} n_1 n_2 n_3 + r_{\tilde{\mathcal{X}}}(n_1+n_2)n_3^2)$. Updating $\bar{\mathcal{L}}_{t+1}$ requires $\mathcal{O}(n_1 n_2 n_3^2 + r_{\tilde{\mathcal{X}}} n_1 n_2 n_3)$. Both $\mathcal{E}_{t+1}$ and $\mathcal{N}_{t+1}$ incur $\mathcal{O}((r_{\tilde{\mathcal{X}}}+1)\,n_1 n_2 n_3)$. Overall, the total computational complexity is $\mathcal{O}(n_1 n_2 n_3^2 + n_3[r_{\tilde{\mathcal{X}}} n_2^2 +(r_{\tilde{\mathcal{X}}}+1) n_1 n_2]).$

Besides, the space complexity of each iteration in Algorithm \ref{algorithm：1} is $\mathcal{O}(n_1 n_2 n_3)$, which is much lower than methods based on T-SVD due to $r_{\tilde{\mathcal{X}}} \ll \min(n_1, n_2)$.

\begin{figure}[t]
    \centering
    \subfigure[$\mathcal{Z}(:,:,1)$]{
        \label{RELF2-a}
        \includegraphics[width=0.20\textwidth]{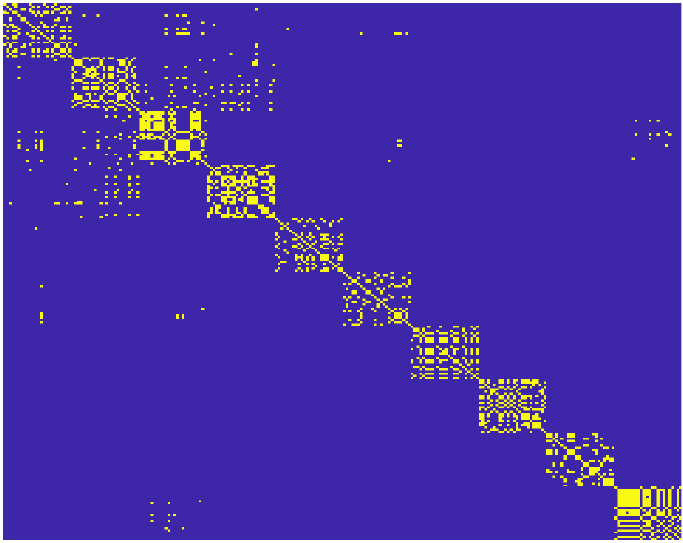}
    } 
    \subfigure[$\mathcal{Z}(:,:,2)$]{
        \label{RELF2-b}
        \includegraphics[width=0.20\textwidth]{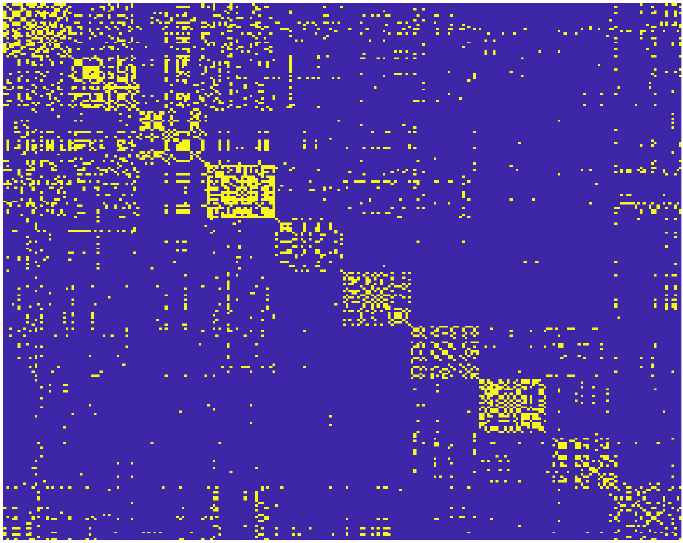}
    }\\
    \subfigure[Traditional weight strategy]{
        \label{RELF2-c}
        \includegraphics[width=0.20\textwidth]{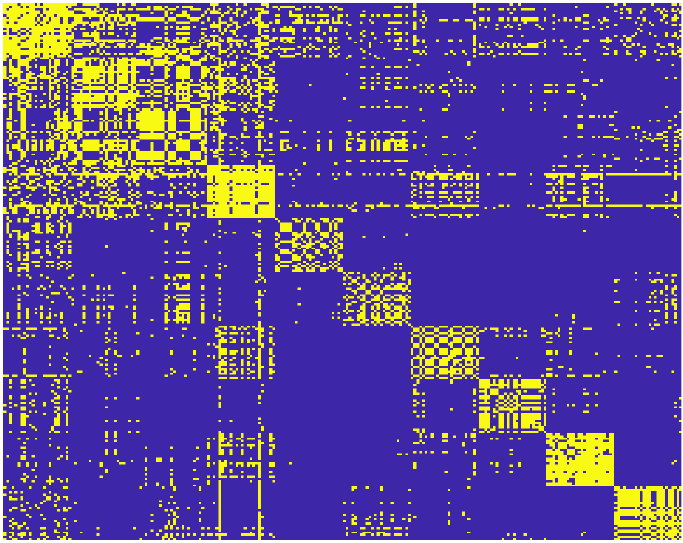}
    } 
    \subfigure[Our weight strategy]{
        \label{RELF2-d}
        \includegraphics[width=0.20\textwidth]{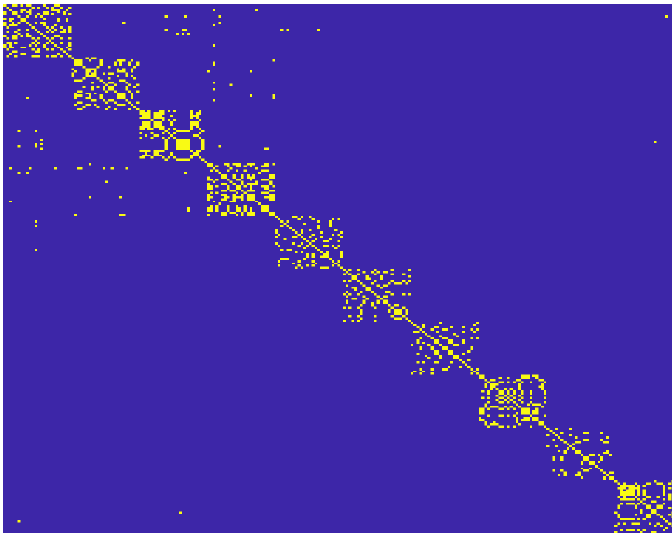}
    }
    \caption{Visualization on the Umist dataset, where (a)-(b) are the representative frontal slices of the coefficient tensor, (c) is the affinity matrix from simple averaging over all slices, and (d) is the one via our proposed diagonal-ratio-based weighted fusion.}
    \label{RELF2}
\end{figure}

\begin{table*}[t]
\centering
\caption{ACC (\%) results of all compared methods under eight benchmark datasets, where the best two results \\are highlighted in \textbf{bold} and \underline{underlined}.}
\label{tab:results1}
\renewcommand{\arraystretch}{1.3}
\begin{tabular}{|c|c|c|c|c|c|c|c|c|}
\hline
Methods & ORL & Umist  & Extended YaleB  & UCSD & Notting-Hill  & MSTAR & GTSRB   & MNIST  \\
\hline
\hline
LLRR   (2011)      & 69.10(±{2.78})  & 72.66(±2.29)  & 78.52(±3.76)  & 92.17(±2.71)  & 54.44(±1.37)  & 62.99(±3.02)  & 53.81(±3.35) & 47.53(±0.60)  \\
\hline
TLRR   (2021)      & 59.03(±2.78)  & 53.77(±3.35)  & 80.68(±3.55)  & \underline{96.98}(±\underline{3.50})  & 53.77(±3.35)  & 72.84(±3.35)  & \underline{54.86}(±\underline{3.10}) & 33.44(±0.99) \\
\hline
ETLRR  (2022)      & 58.24(±2.69)  & 57.61(±3.53)  & 76.99(±3.47)  & 72.02(±4.98) & 52.52(±2.70)  & 75.82(±3.54)  & 50.55(±0.60) & 30.37(±0.72) \\
\hline
IRFLLRR (2023)     & 68.96(±2.33)  & \underline{74.47}(±\underline{2.84})  & 78.56(±3.22)  & 88.14(±3.63)  & \underline{59.33}(±\underline{4.33})& 80.82(±4.69)  & 53.25(±0.04) &\underline{58.35}(±\underline{0.71}) \\
\hline
TS1-LLRR (2024)      & \underline{70.26}(±\underline{2.09})  & 73.64(±3.44)  & 71.08(±3.39)  & 88.01(±3.17) & 56.61(±3.48)  & \underline{84.47}(±\underline{4.20}) & 51.75(±7.19) & 48.27(±1.16)\\
\hline
ESCFE  (2024)   & 67.32(±1.71)  & 71.75(±2.50)  & \underline{82.77}(±\underline{3.90})  & 87.72(±0.00)  &  52.31(±4.91) & 80.48(±2.48)  & 53.17(±0.01)&58.35(±0.89)\\
\hline
BTLRR (2025)    & 53.53(±2.92) & 47.37(±2.99) & 79.94(±3.54) &95.45(±3.49)  &30.16(±0.26) 
 & 70.46(±3.44) & 42.78(±1.30) & 37.65(±1.67) \\
\hline 
\rowcolor{gray!30}  TBTLRR (Our)   & \textbf{72.50}(±\textbf{3.47})  & \textbf{79.92}(±\textbf{2.60})  & \textbf{84.42}(±\textbf{3.41})  & \textbf{100}(±\textbf{0.00})  & \textbf{65.10}(±\textbf{6.72}) & \textbf{85.71}(±\textbf{4.16}) & \textbf{58.04}(±\textbf{1.21})& \textbf{60.08}(±\textbf{1.01})\\
\hline
\end{tabular}
\end{table*}

\begin{table*}[!t]
\centering
\caption{NMI (\%) results of all compared methods under eight benchmark datasets, where the best two results \\are highlighted in \textbf{bold} and \underline{underlined}.}
\label{tab:results2}
\renewcommand{\arraystretch}{1.3}
\begin{tabular}{|c|c|c|c|c|c|c|c|c|}
\hline
Methods & ORL & Umist  & Extended YaleB  & UCSD & Notting-Hill  & MSTAR & GTSRB   & MNIST  \\
\hline
\hline

LLRR   (2011)      & 82.95(±0.94)  &  85.26(±1.18)  & 88.19(±1.49)  & 94.33(±1.79)  & 31.34(±3.06)  & 72.21(±1.29)  & 37.91(±2.65)& 43.30(±0.46) \\
\hline
TLRR   (2021)        & 74.97(±1.41)  & 69.01(±2.88)  & 89.31(±1.52)  &   \underline{97.12}(±\underline{2.86})& 69.01(±2.88)  & 68.10(±1.18)  & 34.41(±1.20) & 24.57(±0.61)  \\
\hline
ETLRR  (2022)      & 75.45(±1.50)  & 70.29(±1.84)  & 87.54(±1.52)  &  78.29(±2.26)& 34.43(±2.70)  & 69.88(±0.65)  & 31.69(±0.50)& 21.84(±0.80) \\
\hline
IRFLLRR (2023)     & 83.16(±0.82) & \textbf{85.90}(±\textbf{1.38})  & 88.69(±1.41)  & 90.93(±2.78)  & 40.59(±3.69)  & 80.94(±2.45)  & 31.19(±0.05)& \underline{54.97}(±\underline{0.32}) \\
\hline
TS1-LLRR (2024)     & \underline{83.80}(±\underline{0.95})  & 83.52(±1.75)  & 84.03(±1.79) & 88.76(±3.23) & 34.31(±3.38)  & \underline{88.19}(±\underline{2.43})  & 23.93(±1.19)& 44.63(±0.65) \\
\hline
ESCFE  (2024)    & 81.34(±0.96)  & 84.04(±1.75)  & \underline{90.70}(±\underline{1.39})  & 88.95(±0.00)  & 33.17(±4.89)  & \textbf{88.84}(±\textbf{1.65})  & \underline{39.97}(±\underline{0.02})&45.29(±1.41) \\
\hline
BTLRR (2025)    & 70.45(±1.75) & 59.21(±3.11) &89.38(±1.43)  & 96.09(±2.64) & 1.53(±0.42)
 &64.38(±1.11)  &20.47(±0.40) &28.69(±1.08)  \\
\hline
\rowcolor{gray!30} TBTLRR (Our)    & \textbf{85.04}(±\textbf{1.68})  & \underline{85.80}(±\underline{1.42})  & \textbf{93.10}(±\textbf{1.14})  & \textbf{100}(±\textbf{0.00})  & \textbf{53.40}(±\textbf{4.74})   & 83.82(±1.80)  & \textbf{40.18}(±\textbf{0.82})& \textbf{56.18}(±\textbf{0.75})\\
\hline

\end{tabular}
\end{table*}

\section{Experiments}\label{Clustering}

This section demonstrates the superiority of our proposed TBTLRR over state-of-the-art subspace clustering methods, including LLRR\footnote{https://github.com/hli1221}, TLRR\footnote{https://panzhous.github.io}, ETLRR\footnote{https://github.com/shiqiangdu/TLRSR-and-ETLRR}, IRFLLRR\footnote{https://github.com/wangzhi-swu/IRFLLRR}, TS1-LLRR, ESCFE, and BTLRR\footnote{https://github.com/MengDing56}. All experiments are implemented in Matlab 2024b on the computer with R5-7500F, 3.7 GHz CPU, and 32 GB of memory.

Table \ref{tab:data} summarizes the statistical information of selected datasets, including ORL \cite{samaria1994parameterisation}, Umist \cite{graham1998characterising}, Extended YaleB \cite{georghiades2002few}, UCSD \cite{piao2016submodule}, Notting-Hill \cite{zhang2009character}, MSTAR \cite{zheng2023fast}, GTSRB \cite{stallkamp2012man}, and MNIST \cite{lecun1180gradient}. These datasets cover a wide range from face images and object images to digital images. Based on the sample size, they are divided into three groups, where small, medium, and large represent sample sizes less than 1000, between 1000 and
5000, and greater than 5000, respectively.
To ensure consistency and reduce computational complexity, all datasets are normalized to the range $[0, 1]$ before processing.

\subsection{Experimental Setup}

\subsubsection{Parameter Settings}
In the experiments, all parameters are carefully chosen based on the recommendations provided in the original references. 
For our proposed TBTLRR, two key parameters are involved, i.e., $\lambda$ and $\beta$. A grid search strategy is applied to find the best parameters, where both $\lambda$ and $\beta$ are searched within the range $[10^{-5}, 10^3]$. 
Note that the number of clusters is fixed to the ground-truth number of categories in each dataset. 
In addition, results are reported as mean ± standard deviation over 50 runs of $k$-means with different random initializations.

\subsubsection{Evaluation Metrics}

Two popular metrics are chosen to quantitatively evaluate the clustering performance of the compared methods.
ACC is defined as
\begin{equation}
\begin{aligned}
 \textrm{ACC} = \frac{1}{n} \sum_{i=1}^{n} \delta\bigl(L_{\textrm{pred}}(i), L_{\textrm{true}}(i)\bigr),
\end{aligned}
\end{equation}
where $L_{\textrm{pred}}$ and $L_{\textrm{true}}$ represent the predicted label and the true label, respectively, and $\delta(\cdot, \cdot)$ is the indicator function that satisfies $\delta(x, y)=1$ if $x = y$ and $\delta(x, y)=0$ otherwise. 

NMI is defined as
\begin{equation}
\begin{aligned}
 \textrm{NMI} = \frac{I(L_{\textrm{pred}}, L_{\textrm{true}})}{\max\bigl(H(L_{\textrm{pred}}), H(L_{\textrm{true}})\bigr)},
\end{aligned}
\end{equation}
where $I(\cdot,\cdot)$ denotes the mutual information, which represents the amount of information shared between $L_{\textrm{pred}}$ and $L_{\textrm{true}}$, and $H(\cdot)$ represents the entropy.

\subsection{Weight Strategy}

Once the coefficient tensor $\mathcal{Z} \in \mathbb{R}^{n_2 \times n_2 \times n_3}$ is obtained, the affinity matrix is typically constructed by averaging all frontal slices, as shown below
\begin{equation}\label{matrix}
{\hat{\mathbf{Z}}} = \frac{1}{2n_3} \sum_{i=1}^{n_3} \bigl( |\mathcal{Z}^{(i)}| + | \bigl( \mathcal{Z}^{(i)} \bigr)^{'} |\bigr).
\end{equation}

However, the quality of individual slices $\mathcal{Z}^{(i)}$ varies considerably. Some slices exhibit clear block-diagonal structures, indicating strong intra-cluster similarities, while others are heavily corrupted by noise and contain little meaningful information. It is found in Fig. \ref{fig:tubal-rank}, the first few slices capture most of the signal's energy, while later slices exhibit rapid decay, indicating noise. This imbalance suggests that equal treatment of all slices is suboptimal for affinity matrix construction.

\begin{figure*}[t]
\centering
\subfigure[ORL (ACC)]{
    \includegraphics[width=4.2cm]{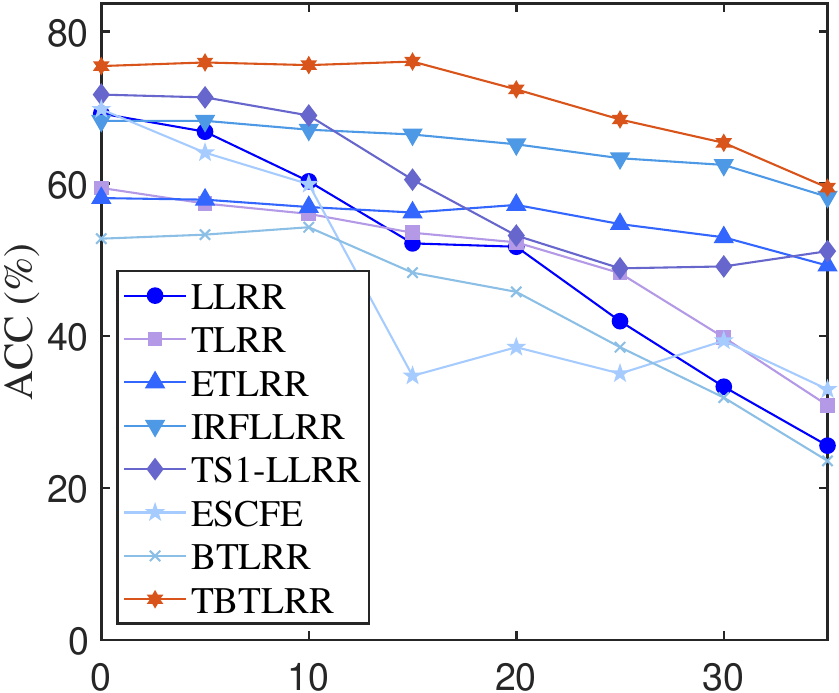}
}\hspace{-1mm}
\subfigure[Umist (ACC)]{
    \includegraphics[width=4.2cm]{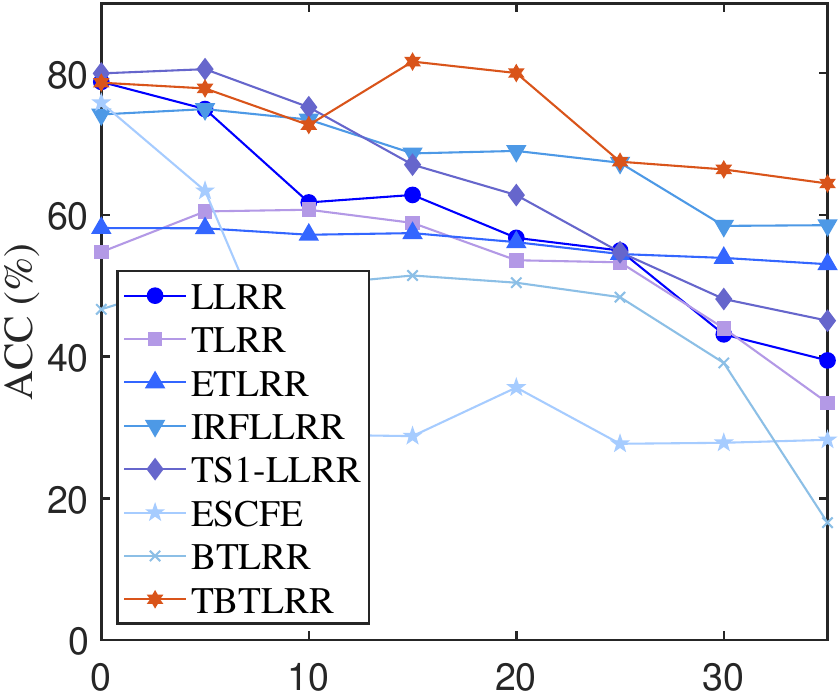}
}\hspace{-1mm}
\subfigure[Extended YaleB (ACC)]{
    \includegraphics[width=4.2cm]{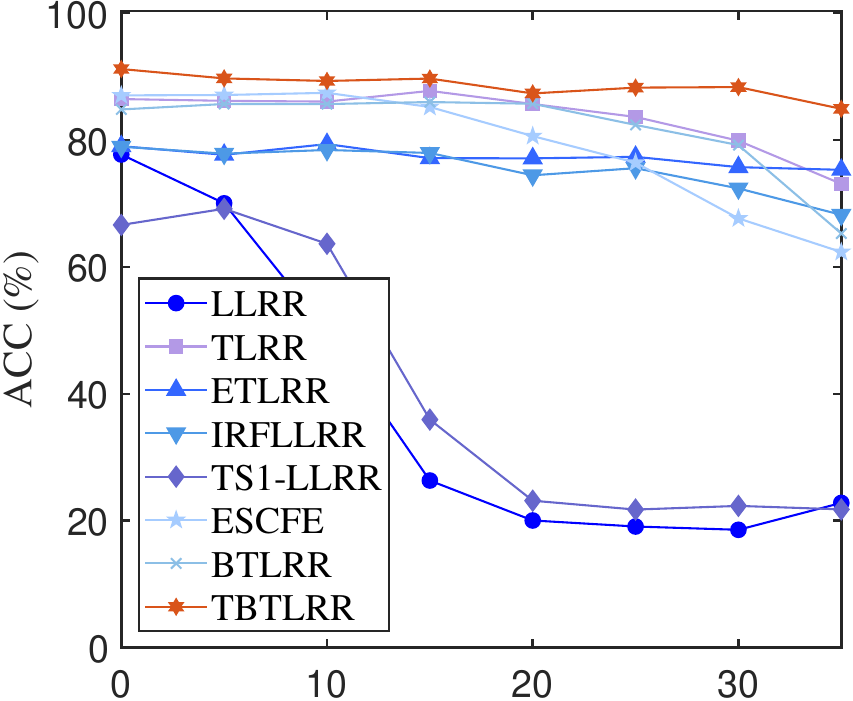}
}\hspace{-1mm}
\subfigure[UCSD (ACC)]{
    \includegraphics[width=4.2cm]{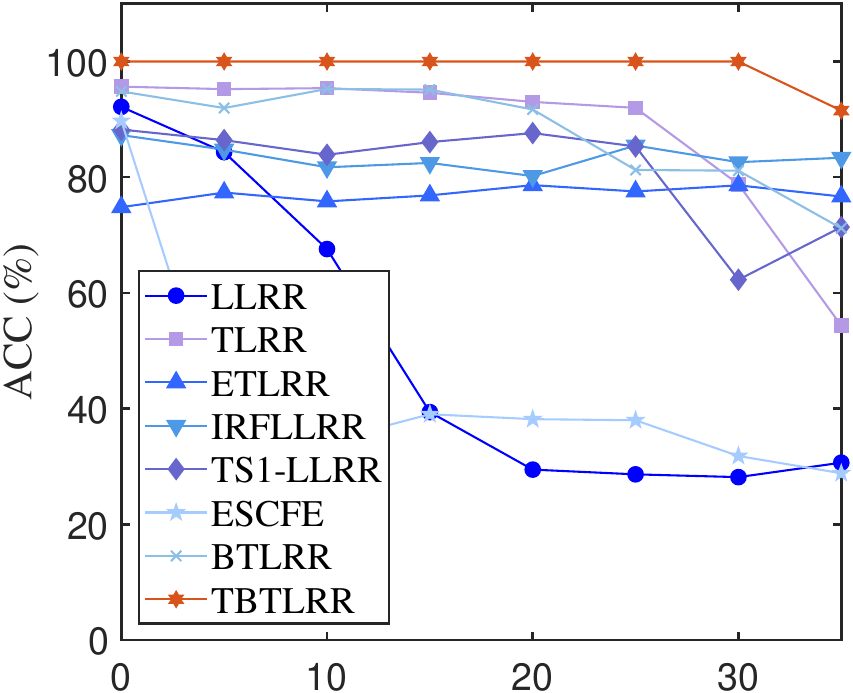}
}
\vspace{0.5cm}  
\subfigure[ORL (NMI)]{
    \includegraphics[width=4.2cm]{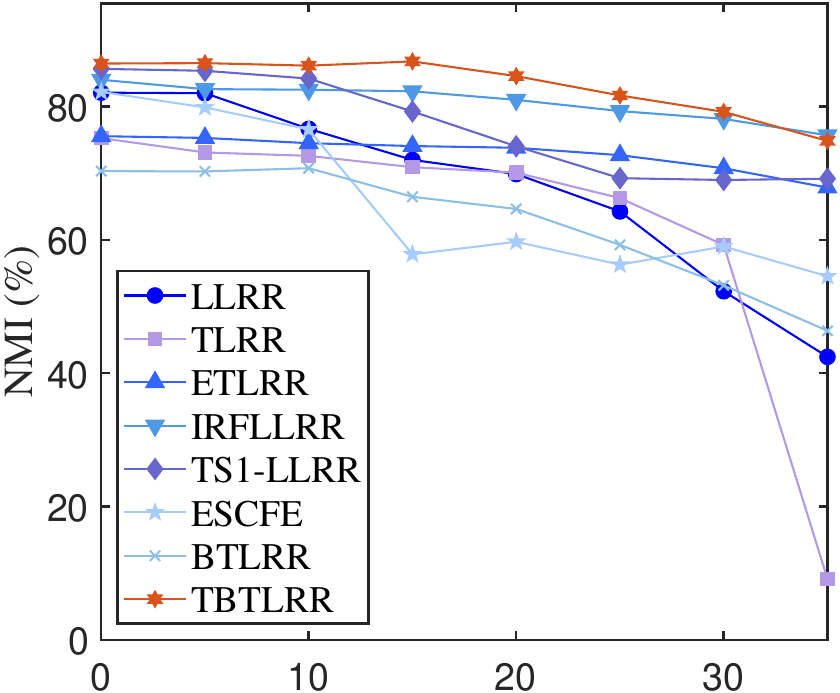}
}\hspace{-1mm}
\subfigure[Umist (NMI)]{
    \includegraphics[width=4.2cm]{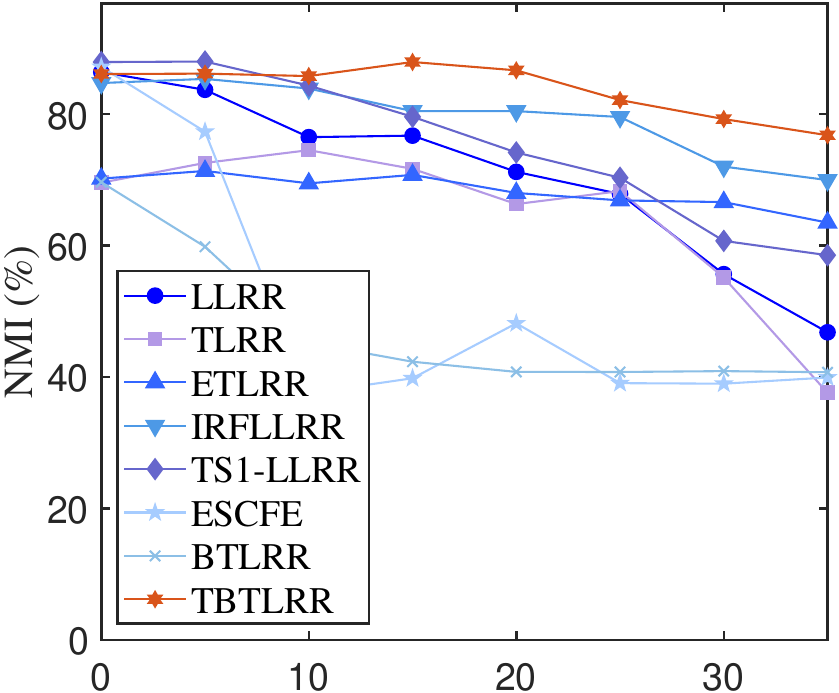}
}\hspace{-1mm}
\subfigure[Extended YaleB (NMI)]{
    \includegraphics[width=4.2cm]{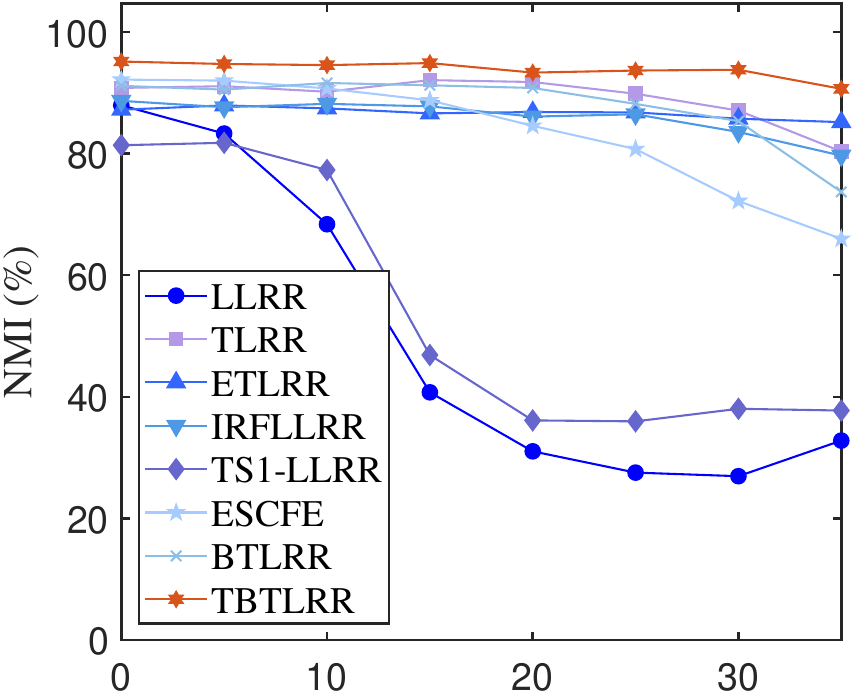}
}\hspace{-1mm}
\subfigure[UCSD (NMI)]{
    \includegraphics[width=4.2cm]{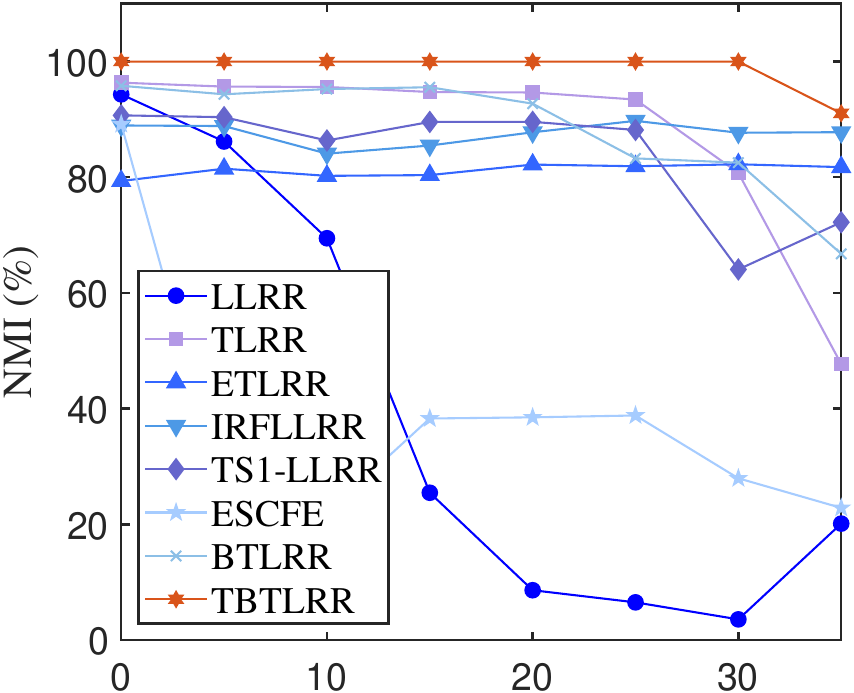}
}
\vspace{-0.5cm}   
\caption{ACC  (\%) and NMI  (\%) of all compared methods on different datasets with varying proportions of sparse noise.}
\label{fig:robust1}
\end{figure*}

\begin{figure*}[t]
\centering
\subfigure[ORL (ACC)]{
    \includegraphics[width=4.2cm]{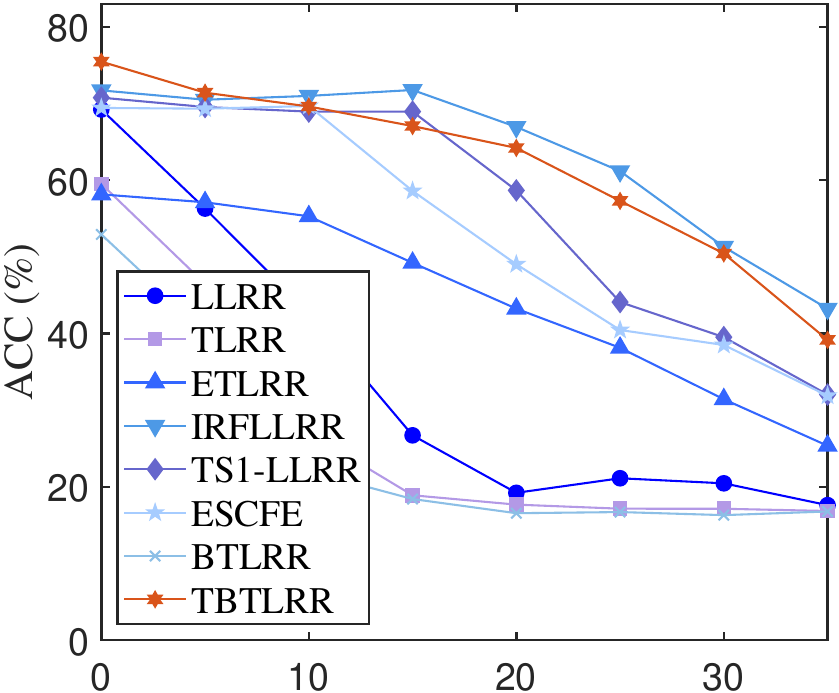}
}\hspace{-1mm}
\subfigure[Umist (ACC)]{
    \includegraphics[width=4.2cm]{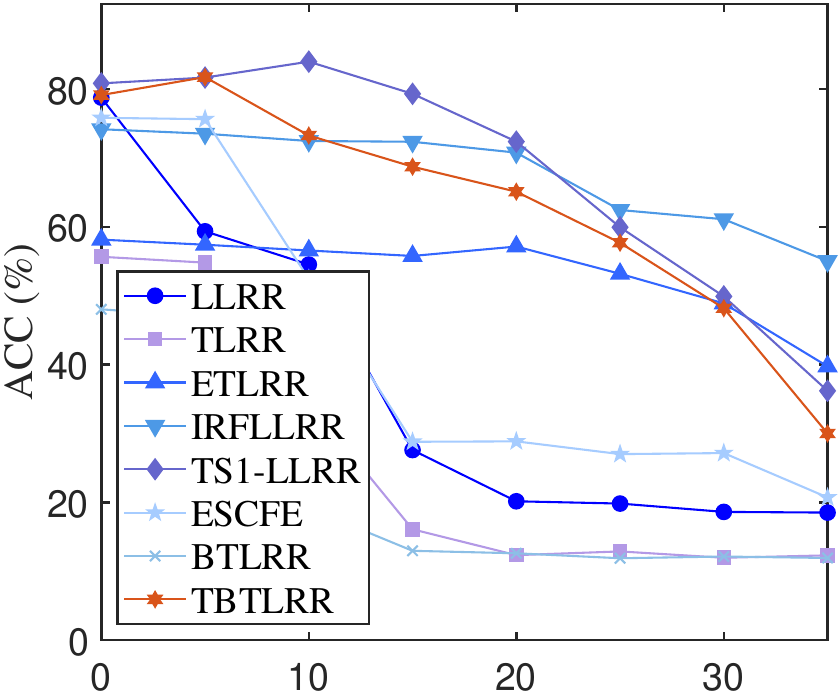}
}\hspace{-1mm}
\subfigure[Extended YaleB (ACC)]{
    \includegraphics[width=4.2cm]{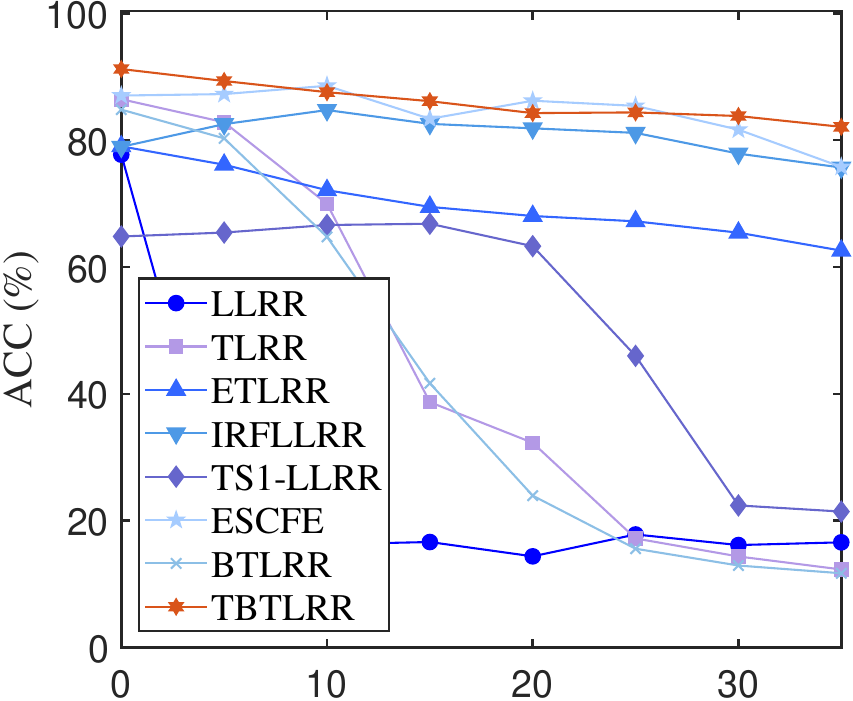}
}\hspace{-1mm}
\subfigure[UCSD (ACC)]{
    \includegraphics[width=4.2cm]{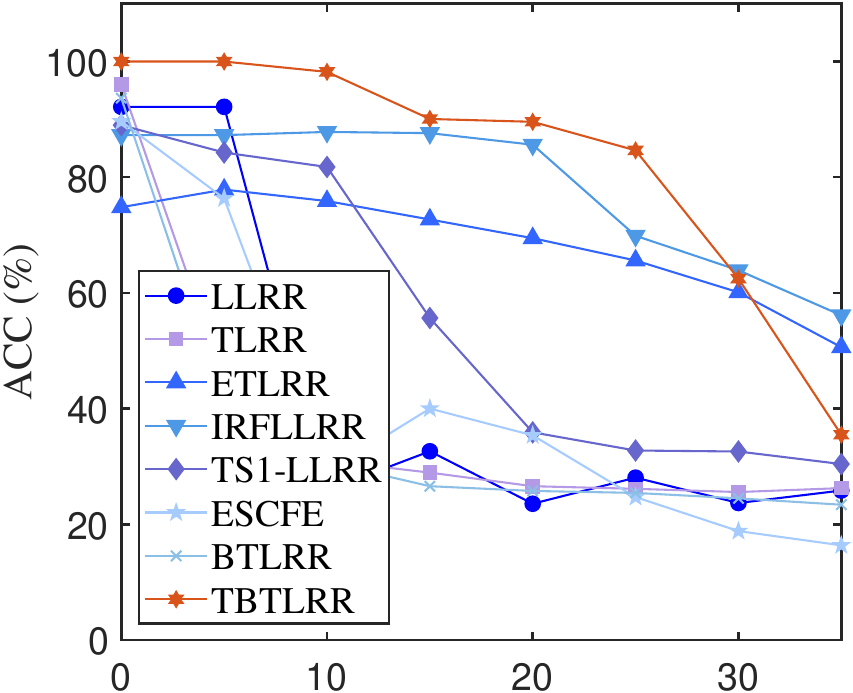}
}
\vspace{0.5cm}  
\subfigure[ORL (NMI)]{
    \includegraphics[width=4.2cm]{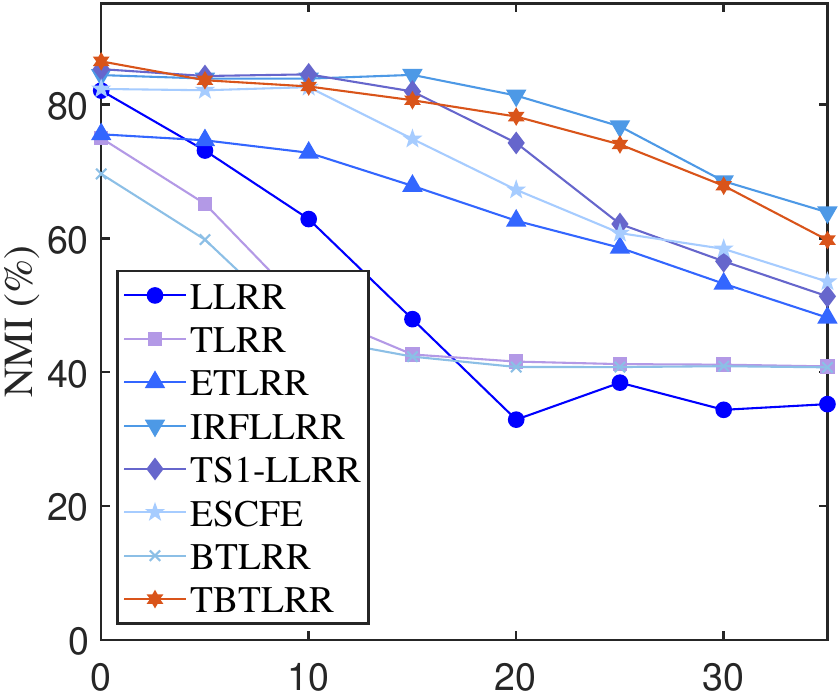}
}\hspace{-1mm}
\subfigure[Umist (NMI)]{
    \includegraphics[width=4.2cm]{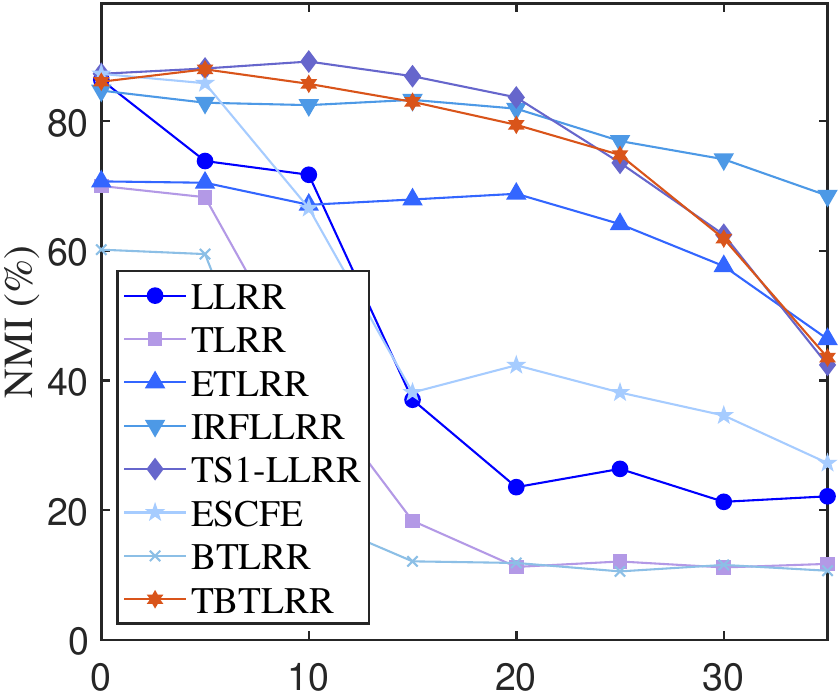}
}\hspace{-1mm}
\subfigure[Extended YaleB (NMI)]{
    \includegraphics[width=4.2cm]{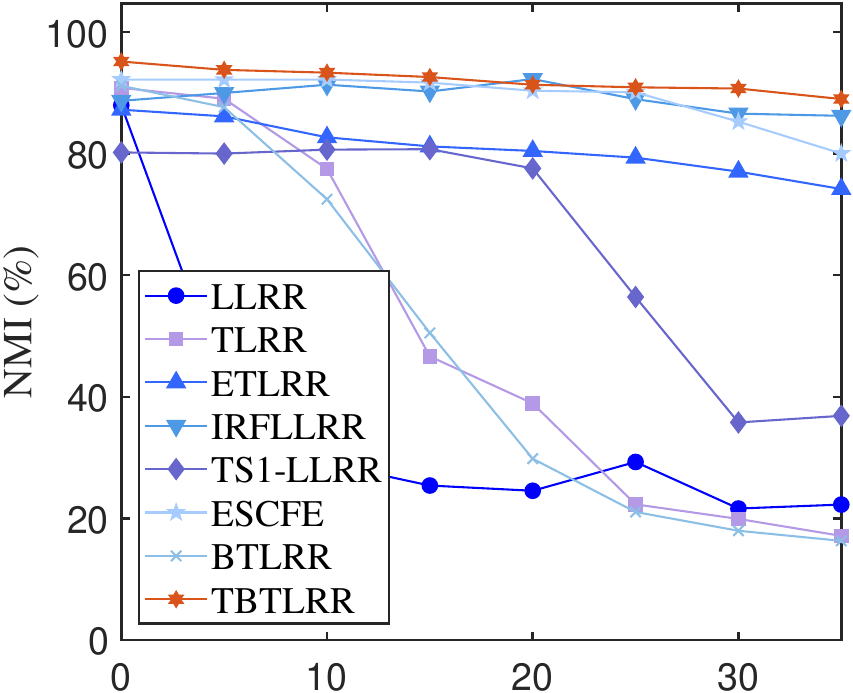}
}\hspace{-1mm}
\subfigure[UCSD (NMI)]{
    \includegraphics[width=4.2cm]{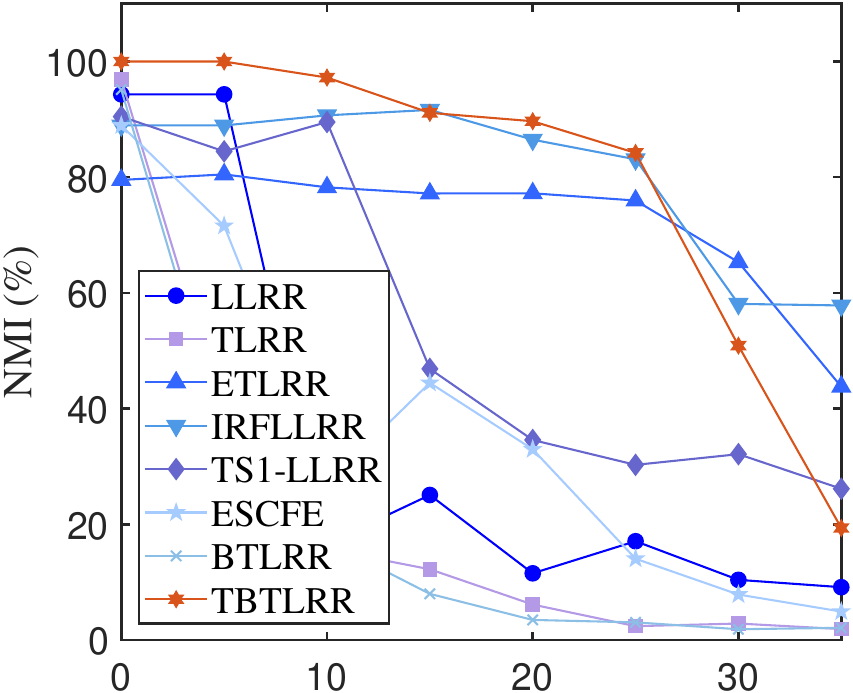}
}
\vspace{-0.5cm}
\caption{ACC  (\%) and NMI  (\%) of all compared methods on different datasets with varying proportions of Gaussian noise.}
\label{fig:robust2}
\end{figure*}

To address this, we introduce a diagonal-ratio-based slice weight strategy. Each slice $\mathcal{Z}^{(i)}$ is assigned a weight based on its diagonal energy ratio, calculated as 
\begin{equation}
 r_i = \frac{\sum_{j=1}^{n_2} |\mathcal{Z}^{(i)}_{jj}|}{\sum_{j=1}^{n_2} \sum_{k=1}^{n_2} |\mathcal{Z}^{(i)}_{jk}| + \epsilon},
\end{equation}
where $\epsilon>0$ prevents division by zero. This ratio reflects the concentration of affinity values along the diagonal, which correlates with the slice's structural quality. The weights are then normalized by
\begin{equation}
 w_i = \frac{r_i}{\sum_{j=1}^{n_3} r_j} \quad \textrm{and} \quad \sum_{i=1}^{n_3} w_i = 1.
\end{equation}

Finally, the affinity matrix is computed as a weighted sum of slices, i.e.,
\begin{equation}
{\hat{\mathbf{Z}} = \sum_{i=1}^{n_3} w_i \mathcal{Z}^{(i)}.}
\end{equation}

Fig. \ref{RELF2-a} shows a slice with a clear block-diagonal structure, whereas Fig. \ref{RELF2-b} presents a noisy slice. 
The resulting affinity matrix in Fig. \ref{RELF2-d} exhibits a clearer block-diagonal structure compared to the simple averaging result in Fig. \ref{RELF2-c}. It validates that our proposed weighting strategy emphasizes well-structured slices and suppresses noisy ones.

\begin{figure*}[t]
\centering
\subfigure[LLRR]{
    \label{a}
    \centering
    \includegraphics[width=3.5cm]{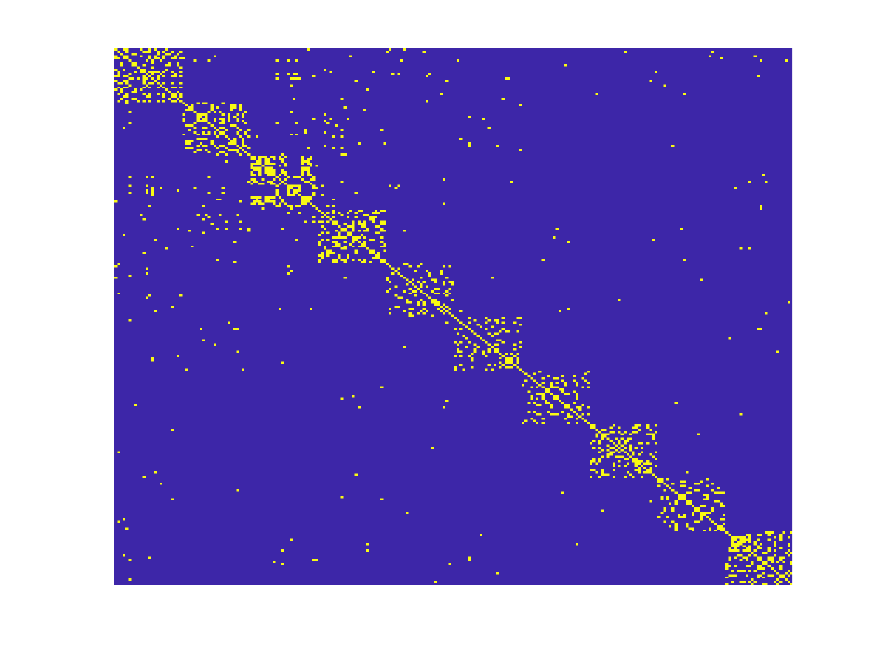}
}
\subfigure[TLRR]{
    \label{b}
    \centering
    \includegraphics[width=3.5cm]{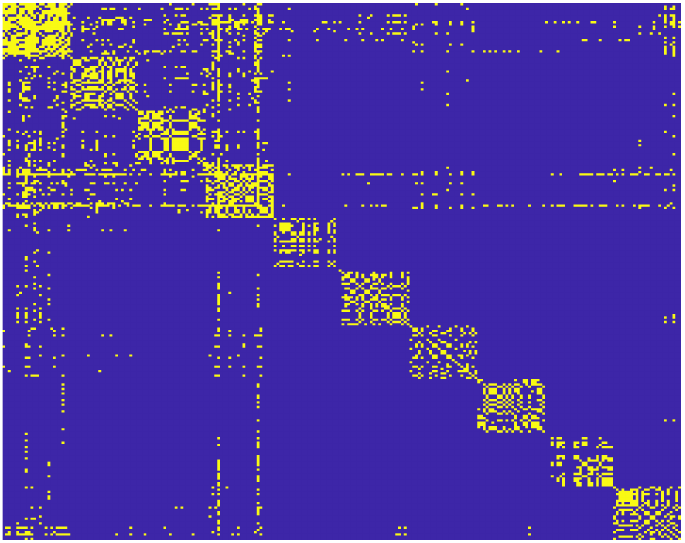}
}
\subfigure[ETLRR]{
    \label{c}
    \centering
    \includegraphics[width=3.5cm]{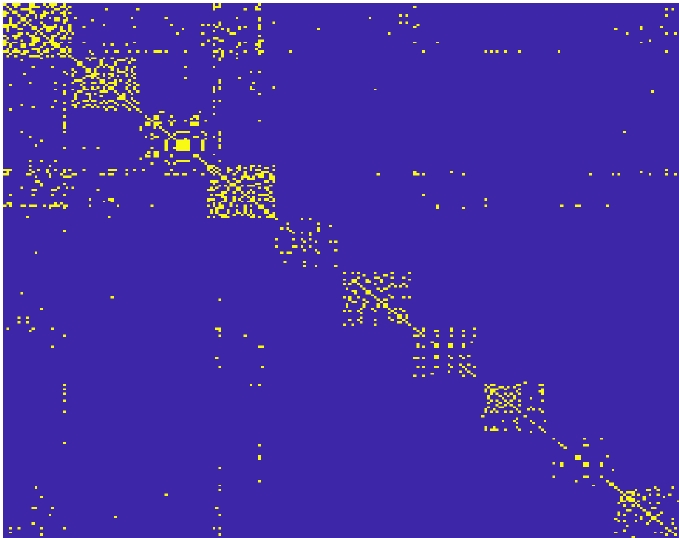}
}
\subfigure[IRFLLRR]{
    \label{d}
    \centering
    \includegraphics[width=3.5cm]{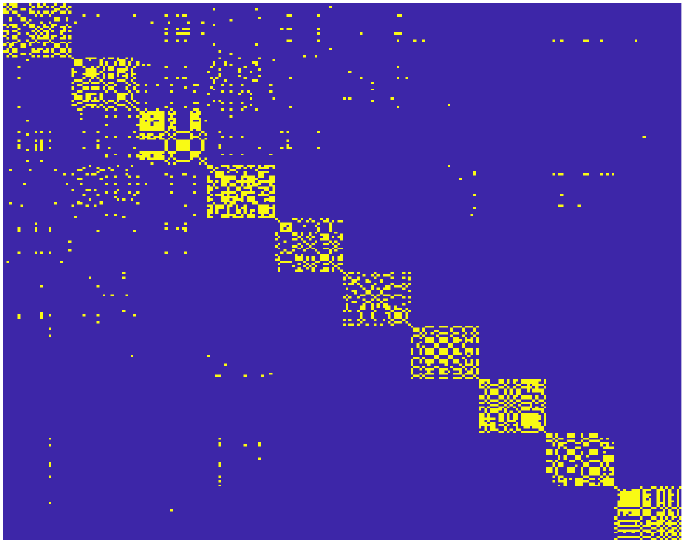}
}
\subfigure[TS1-LLRR]{
    \label{e}
    \centering
    \includegraphics[width=3.5cm]{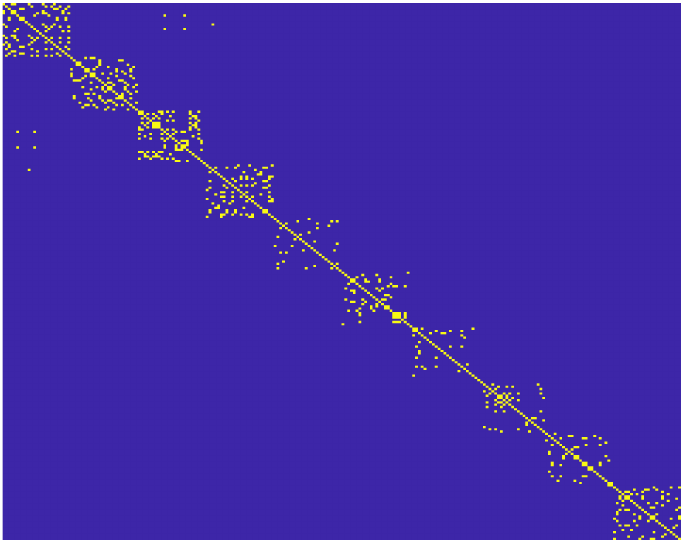}
}
\subfigure[ESCFE]{
    \label{f}
    \centering
    \includegraphics[width=3.5cm]{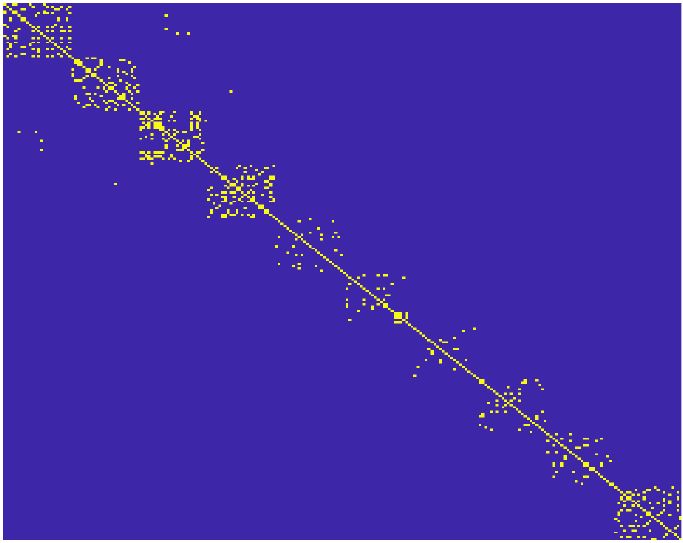}
}
\subfigure[BTLRR]{
    \label{g}
    \centering
    \includegraphics[width=3.5cm]{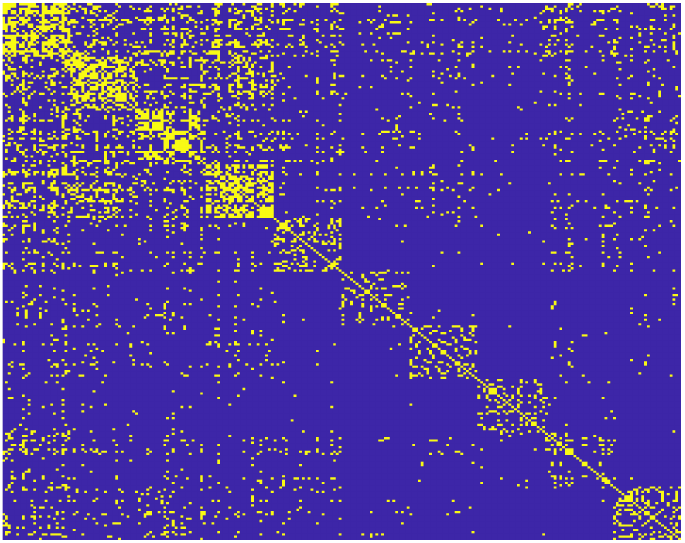}
}
\subfigure[TBTLRR]{
    \label{h}
    \centering
    \includegraphics[width=3.5cm]{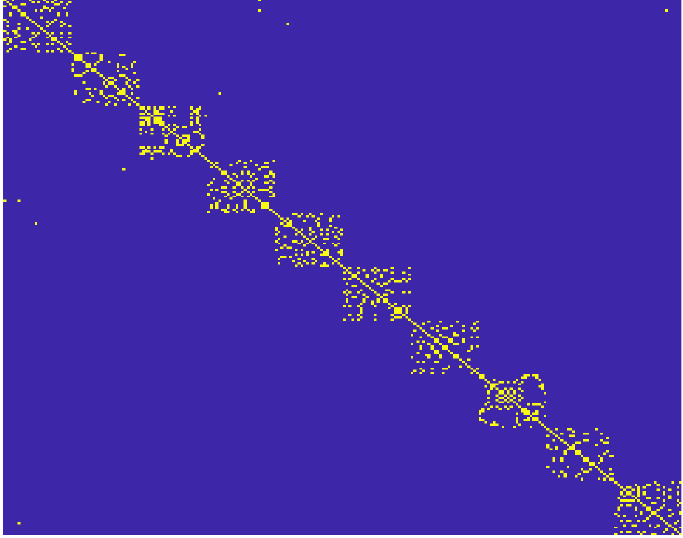}
}
\caption{Visual comparison of the affinity matrices obtained by various methods on the first 10 classes on the Umist dataset with 10\% noise.}
\label{fig:matrix1}
\end{figure*}

\begin{figure*}[t]
\centering
\subfigure[LLRR]{
    \label{a}
    \centering
    \includegraphics[width=3.5cm]{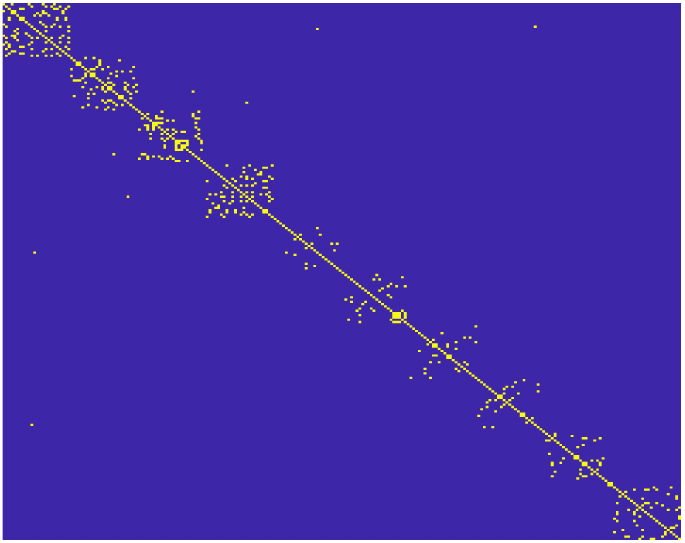}
}
\subfigure[TLRR]{
    \label{b}
    \centering
    \includegraphics[width=3.5cm]{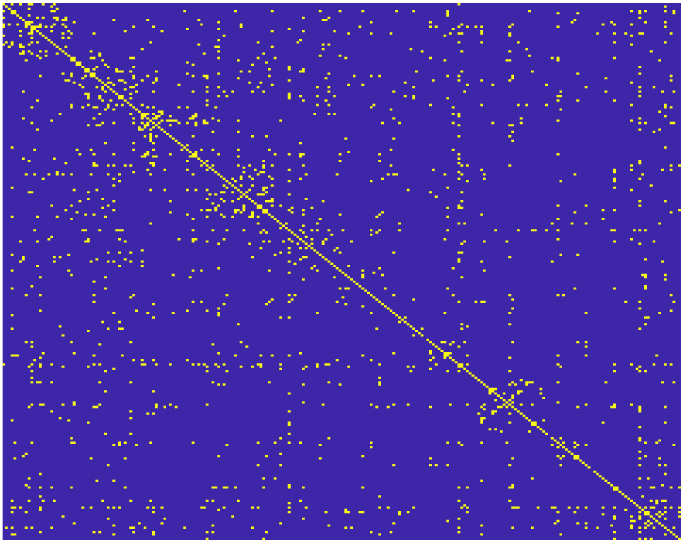}
}
\subfigure[ETLRR]{
    \label{c}
    \centering
    \includegraphics[width=3.5cm]{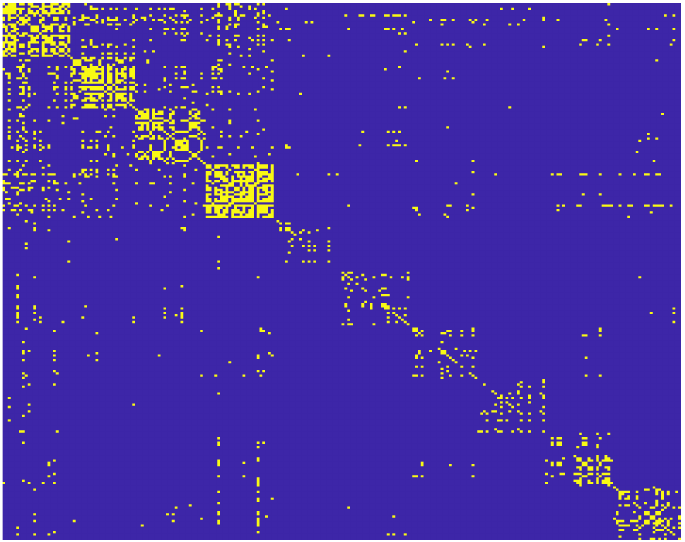}
}
\subfigure[IRFLLRR]{
    \label{d}
    \centering
    \includegraphics[width=3.5cm]{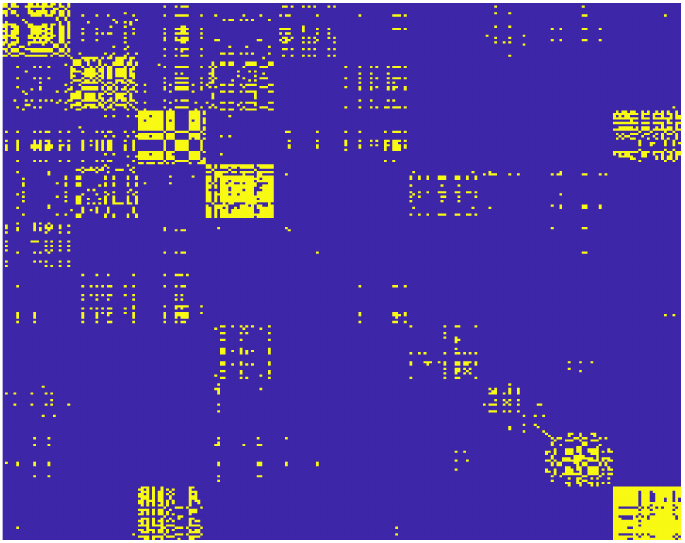}
}
\subfigure[TS1-LLRR]{
    \label{e}
    \centering
    \includegraphics[width=3.5cm]{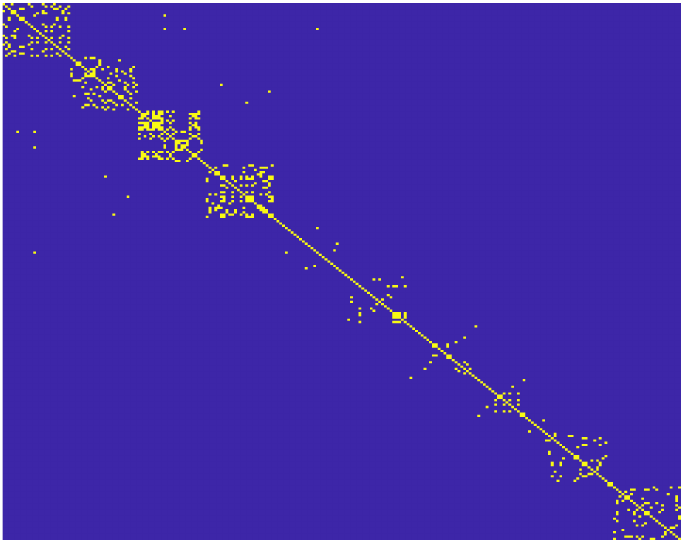}
}
\subfigure[ESCFE]{
    \label{f}
    \centering
    \includegraphics[width=3.5cm]{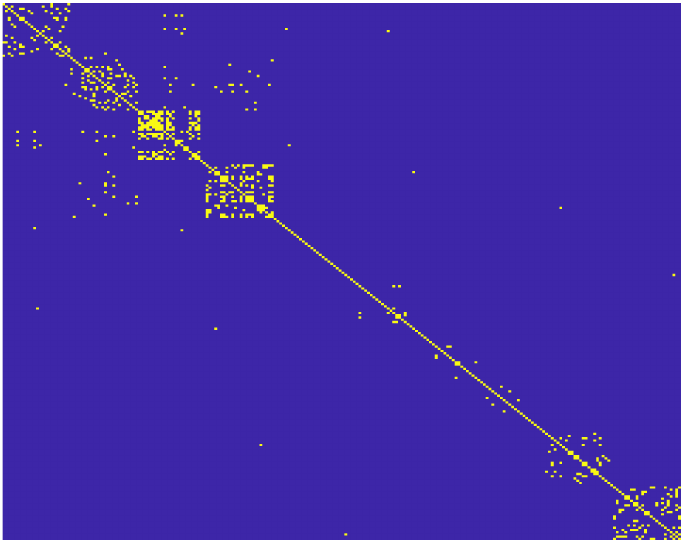}
}
\subfigure[BTLRR]{
    \label{g}
    \centering
    \includegraphics[width=3.5cm]{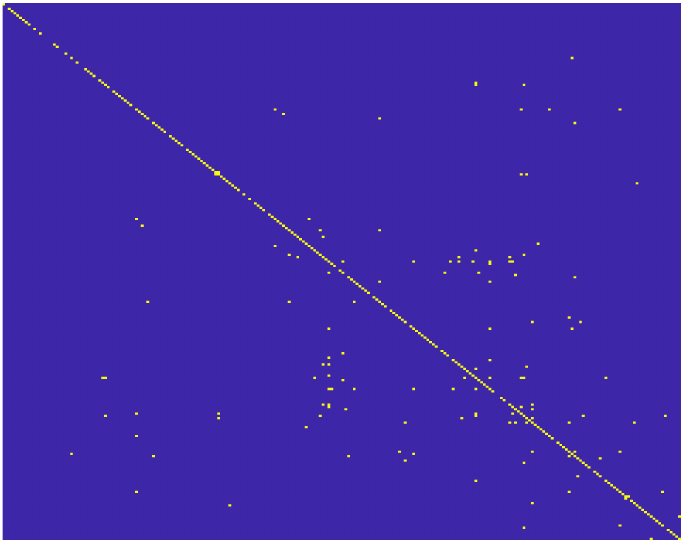}
}
\subfigure[TBTLRR]{
    \label{h}
    \centering
    \includegraphics[width=3.5cm]{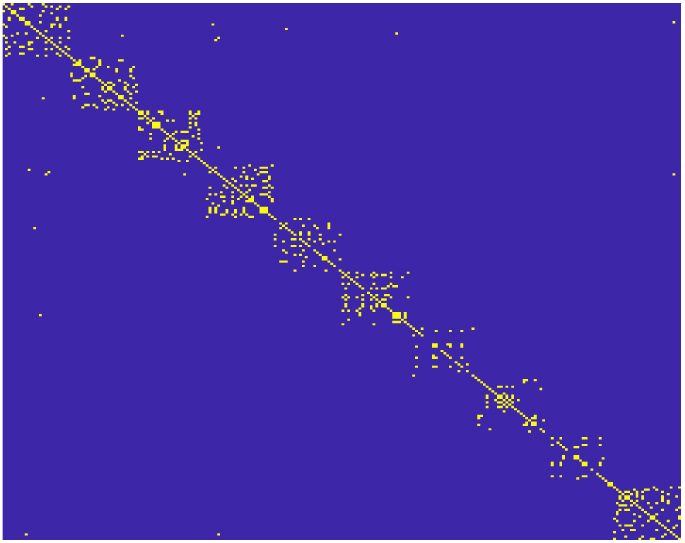}
}
\caption{Visual comparison of the affinity matrices obtained by various methods on the first 10 classes on the Umist dataset with 35\% noise.}
\label{fig:matrix2}
\end{figure*}

\subsection{Numerical Experiments}

The clustering performance of all compared methods is listed in Tables \ref{tab:results1} and \ref{tab:results2}, where the best results in each column are highlighted in \textbf{bold}, while the second best results are \underline{underlined}. The following conclusions can be made.
\begin{itemize}
\item The proposed TBTLRR achieves either the best or the second-best clustering performance across almost all datasets. Notably, on the GTSRB dataset, the ACC of TBTLRR reaches 58.04\%, which is 3.18\% higher than that of the best-performing baseline method. 
\item Among these matrix LRR-based methods, IRFLLRR often achieves the best performance, which can be attributed to the ability to flexibly handle individual singular values through rational weighting. By introducing tensor representation and employing the TTNN under a data-adaptive transform, TBTLRR captures multi-mode correlations and cross-slice consistency,  thereby achieving promising clustering performance.
\item As a representative tensor-based method, TLRR exhibits suboptimal performance on the ORL and Umist datasets, due to its reliance on the TNN based on FFT, which fails to accurately characterize the low-rank structures of these two datasets. In contrast, our proposed TBTLRR  achieves a significant performance improvement on both datasets.
\end{itemize}

\begin{table*}[!h]
    \renewcommand\arraystretch{1.3}
    \caption{Ablation studies of our proposed method, where the best two results are highlighted in \textbf{bold} and \underline{underlined}.}
    \label{ablation}
    \centering
    \begin{tabular}{|l|cc|cc|cc|cc|}
      \hline 
      \multirow{2}{*}{Cases} & 
      \multicolumn{2}{c|}{ORL} & 
      \multicolumn{2}{c|}{Umist} & 
      \multicolumn{2}{c|}{Extended YaleB} &
      \multicolumn{2}{c|}{UCSD} \\
      \cline{2-9}
      & ACC & NMI & ACC & NMI & ACC & NMI & ACC & NMI \\
      \hline\hline
  Case I & 60.23(±2.60) & 77.09(±1.43) & 69.03(±2.06) & 75.80(±1.78)& 81.67(±3.49) & 91.60(±1.29) & \underline{97.25}(±\underline{2.09}) & \underline{97.37}(±\underline{1.30})   \\
      \hline
    Case II & 67.56(±2.96) & 82.17(±1.41) & \underline{79.90}(±\underline{2.25}) & \underline{85.73}(±\underline{1.38}) & 81.66(±3.75) & 91.50(±1.37) & \textbf{100}(±\textbf{0.00}) & \textbf{100}(±\textbf{0.00}) \\
      \hline
   Case III  & \underline{72.25}(±\underline{2.61}) & \underline{84.73}(±\underline{1.28}) & 75.71(±3.06) & 81.20(±1.74)& \underline{82.59}(±\underline{2.83})  & \underline{91.91}(±\underline{1.37}) &\textbf{100}(±\textbf{0.00}) & \textbf{100}(±\textbf{0.00})  \\
      \hline
 \rowcolor{gray!30}    Case IV  & \textbf{72.50}(±\textbf{3.47}) & \textbf{85.04}(±\textbf{1.68}) & \textbf{79.92}(±\textbf{2.60}) & \textbf{85.80}(±\textbf{1.42}) & \textbf{84.42}(±\textbf{3.41}) & \textbf{93.10}(±\textbf{1.14}) & \textbf{100}(±\textbf{0.00}) & \textbf{100}(±\textbf{0.00}) \\
      \hline
    \end{tabular}
\end{table*}

\subsection{Robustness Verification}
This section verifies the robustness on four datasets, i.e., ORL, Umist, Extended YaleB, and UCSD. Fig. \ref{fig:robust1} and Fig. \ref{fig:robust2} show the clustering performance of all comparative methods with varying levels of sparse and Gaussian noise (0\%-35\%).

Although the performance of TBTLRR also declines, the extent of this degradation is relatively minimal compared to the other methods. For instance, on the UCSD dataset with sparse noise, the clustering performance of TBTLRR remains almost unaffected by the noise. On the Extended YaleB dataset with 35\% Gaussian noise, TBTLRR achieves an improvement of at least 2\% in clustering performance.

Furthermore, the affinity matrices obtained by eight methods with 10\% and 35\% sparse noise to the first 10 classes on the Umist dataset are shown in Figs. \ref{fig:matrix1} and \ref{fig:matrix2}, respectively. Our proposed TBTLRR maintains a clear block-diagonal structure even under heavy corruptions. This robustness can be attributed to the data-adaptive transform that enhances low-rankness and two regularization terms that filters out noise.

\begin{figure*}[t]
\hspace{-1cm}
\centering
\subfigure[ORL (ACC)]{
    \label{a}
    \centering
    \includegraphics[width=3.8cm]{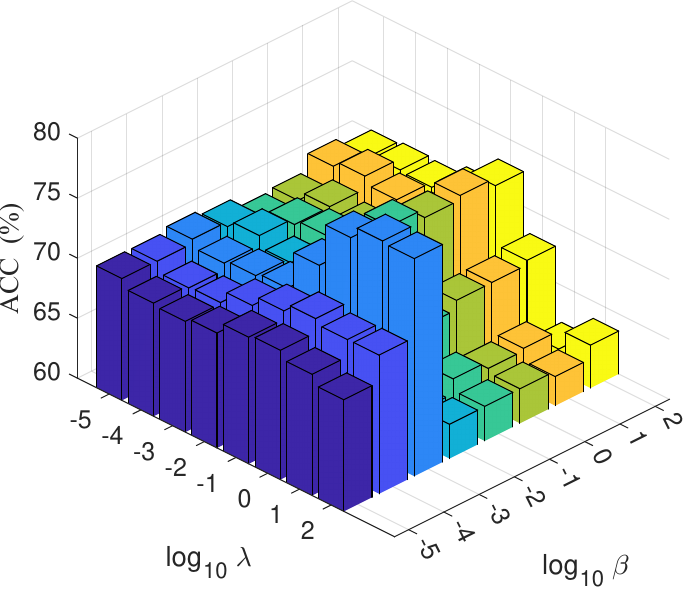}
}\hspace{-0mm}
\subfigure[Umist (ACC)]{
    \label{b}
    \centering
    \includegraphics[width=3.8cm]{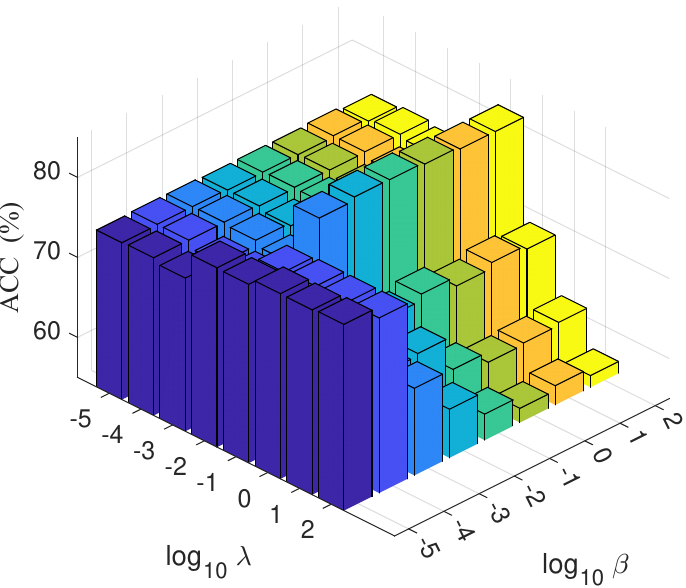}
}\hspace{-0mm}
\subfigure[Extended YaleB  (ACC)]{
    \label{d}
    \centering
    \includegraphics[width=3.8cm]{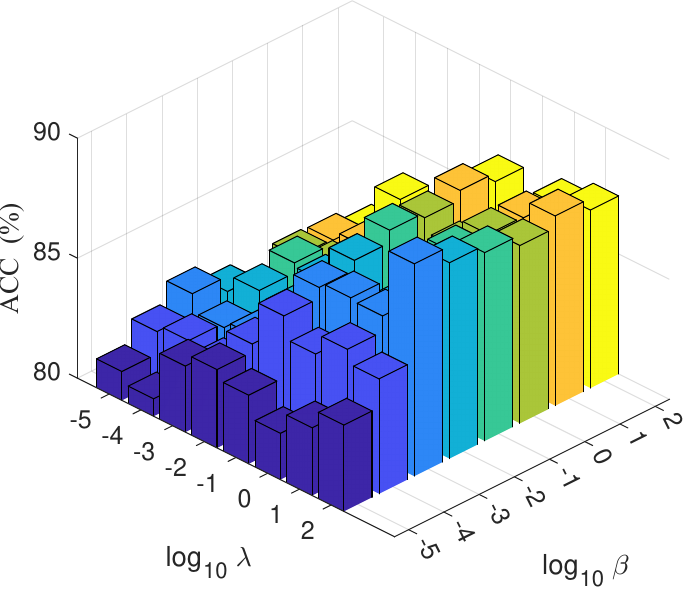}
}\hspace{-0mm}
\subfigure[UCSD  (ACC)]{
    \label{c}
    \centering
    \includegraphics[width=3.8cm]{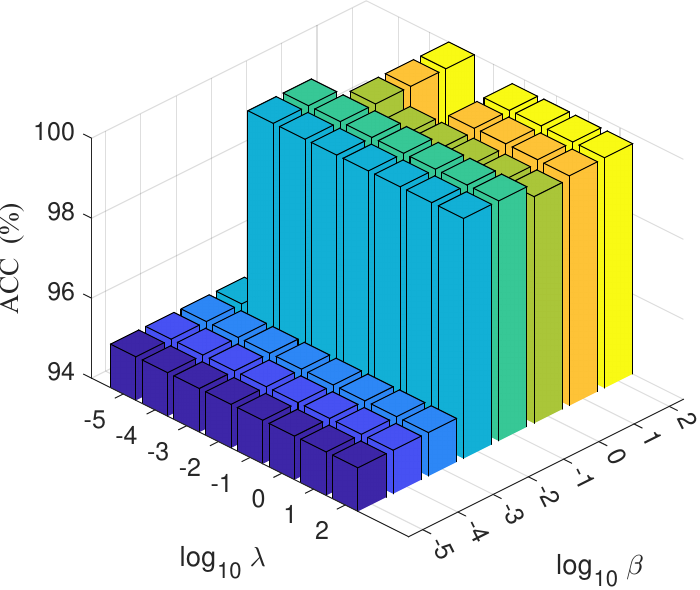}
}\hspace{-0mm}

\hspace{-1cm}
\subfigure[ORL (NMI)]{
    \label{e}
    \centering
    \includegraphics[width=3.8cm]{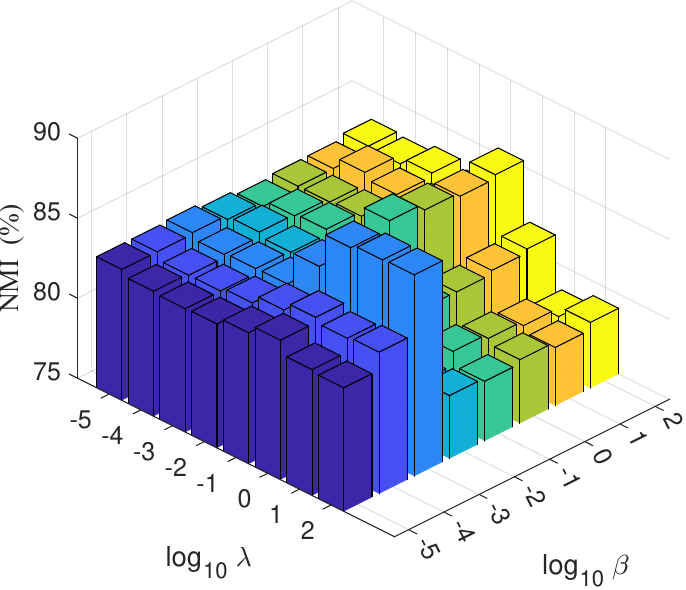}
}\hspace{-0mm}
\subfigure[Umist (NMI)]{
    \label{f}
    \centering
    \includegraphics[width=3.8cm]{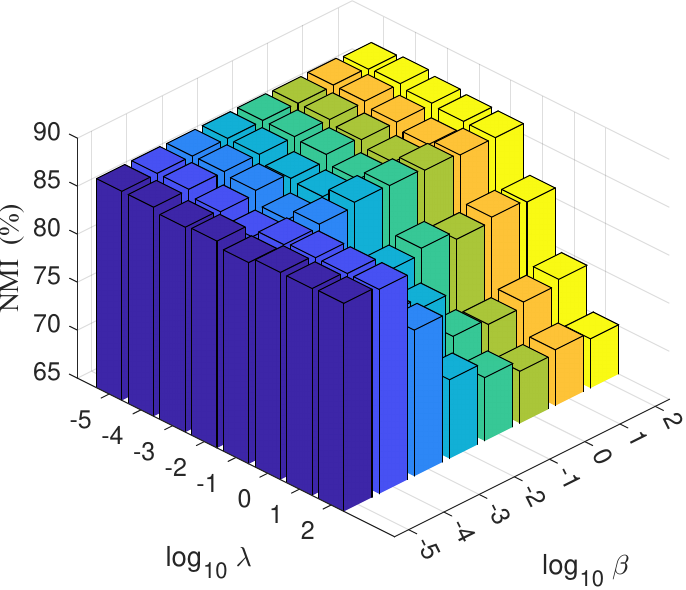}
}\hspace{-0mm}
\subfigure[Extended YaleB  (NMI)]{
    \label{h}
    \centering
    \includegraphics[width=3.8cm]{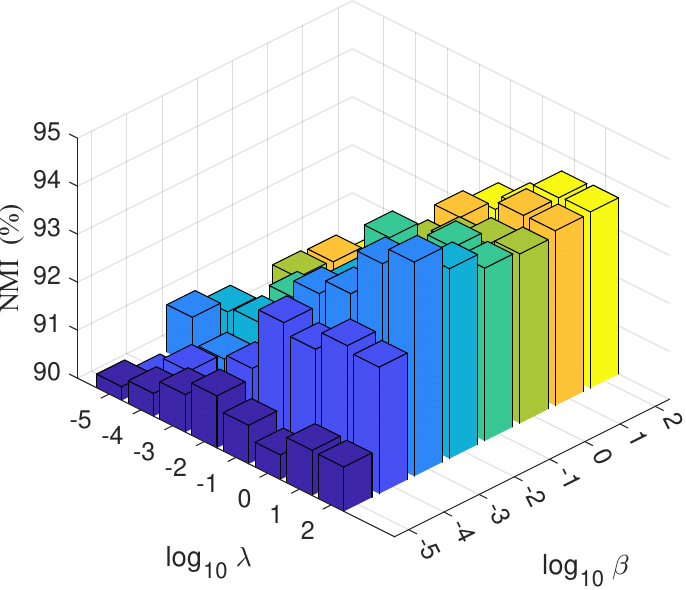}
}\hspace{-0mm}
\subfigure[UCSD (NMI)]{
    \label{g}
    \centering
    \includegraphics[width=3.8cm]{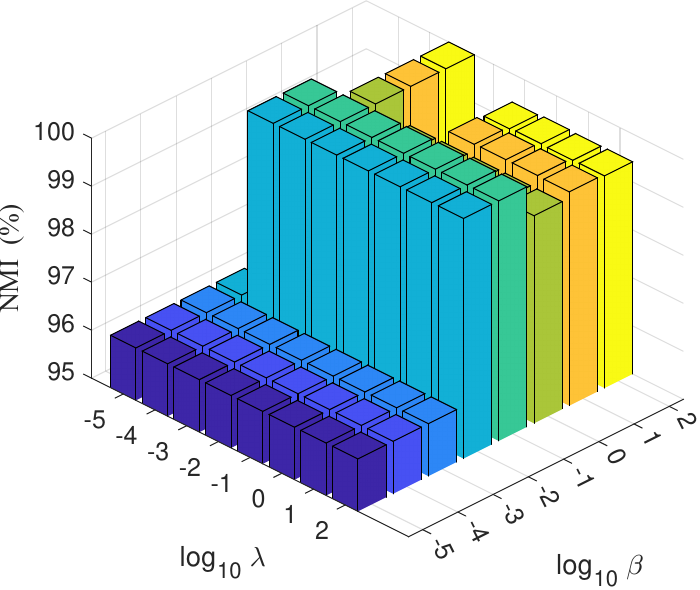}
}\hspace{-0mm}
\caption{Effects of $\lambda$ and $\beta$ on four real-world datasets, where (a)-(d) are the ACC results and (e)-(h) are the NMI results.}
\label{Para}
\end{figure*}

\begin{figure*}[t]
\hspace{-0.5cm}
\centering
\subfigure[ORL (ACC)]{
    \label{a}
    \centering
    \includegraphics[width=3.9cm]{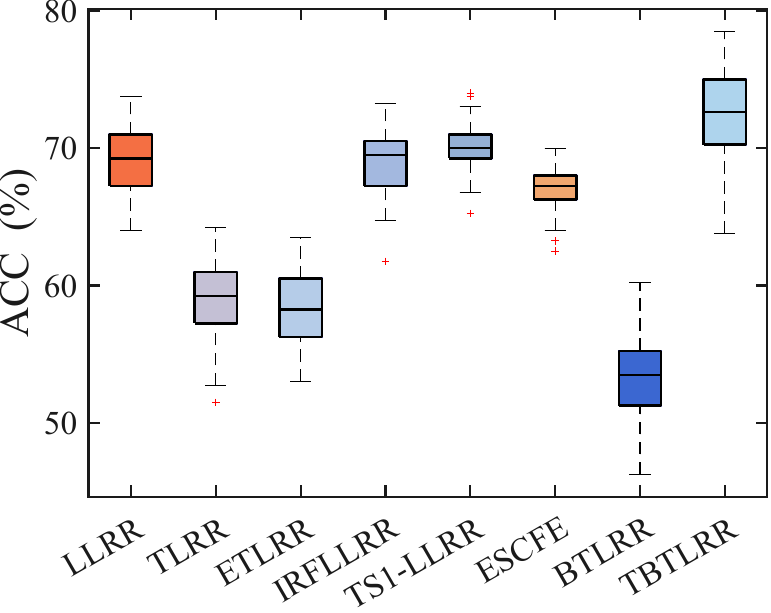}
}\hspace{0mm}
\subfigure[Umist (ACC)]{
    \label{b}
    \centering
    \includegraphics[width=3.9cm]{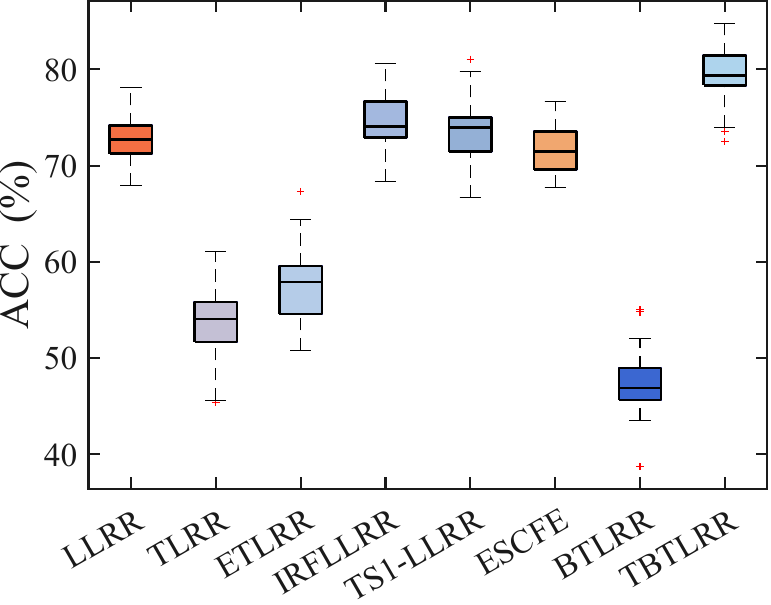}
}\hspace{0mm}
\subfigure[Extended YaleB (ACC)]{
    \label{c}
    \centering
    \includegraphics[width=3.9cm]{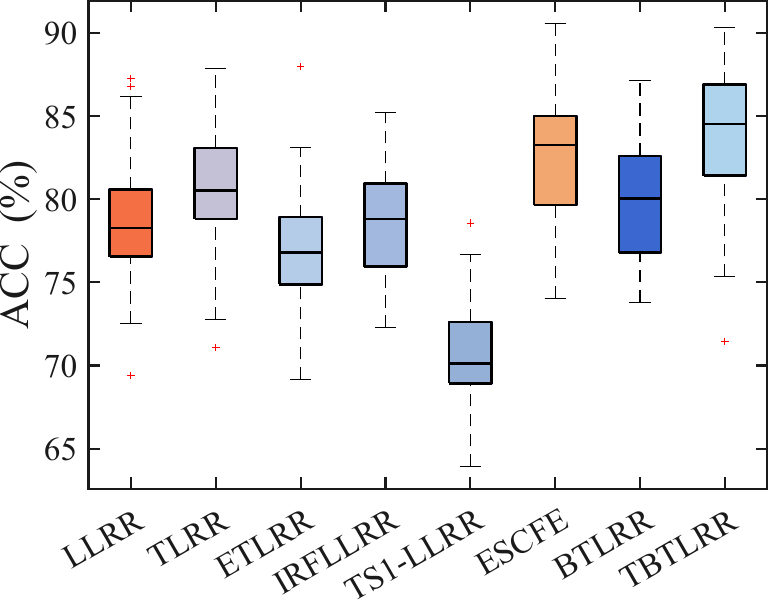}
}\hspace{0mm}
\subfigure[UCSD (ACC)]{
    \label{c}
    \centering
    \includegraphics[width=3.9cm]{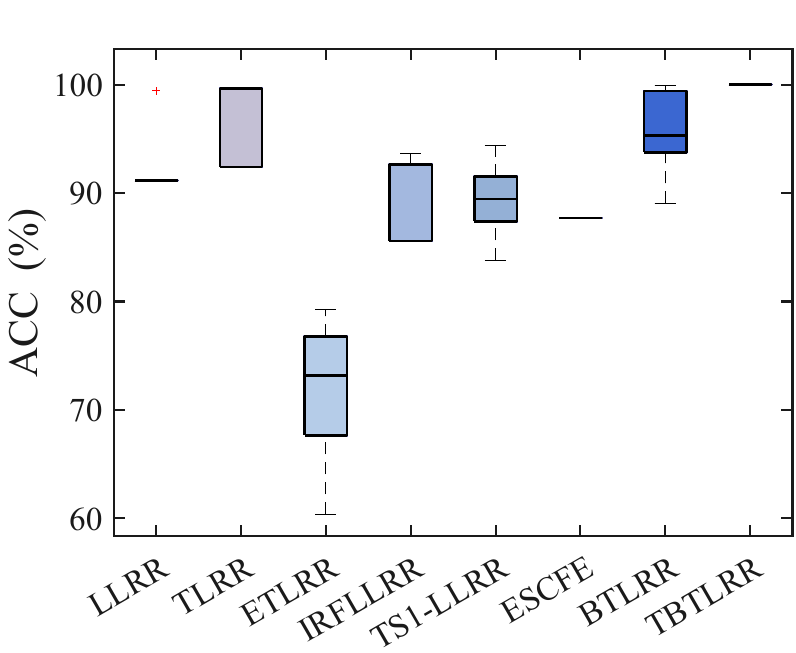}
}\hspace{0mm}

\hspace{-0.5cm}
\subfigure[ORL (NMI)]{
    \label{e}
    \centering
    \includegraphics[width=3.9cm]{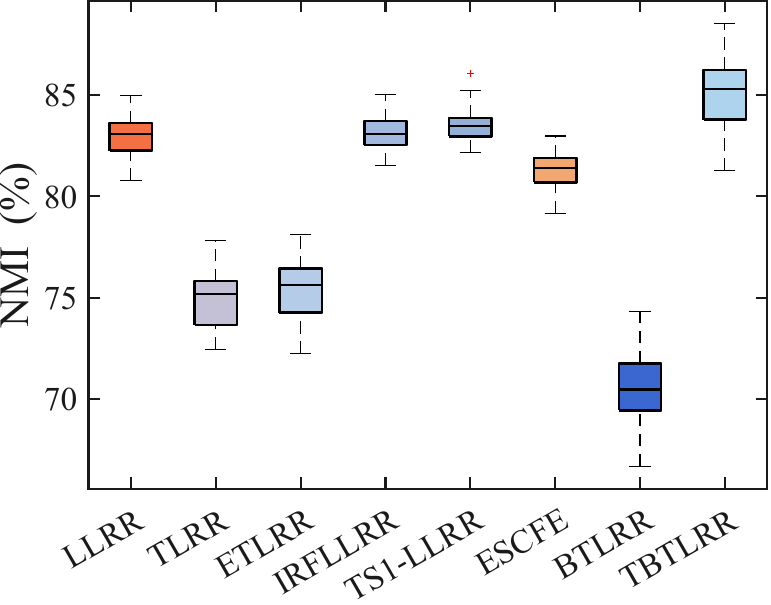}
}\hspace{0mm}
\subfigure[Umist (NMI)]{
    \label{f}
    \centering
    \includegraphics[width=3.9cm]{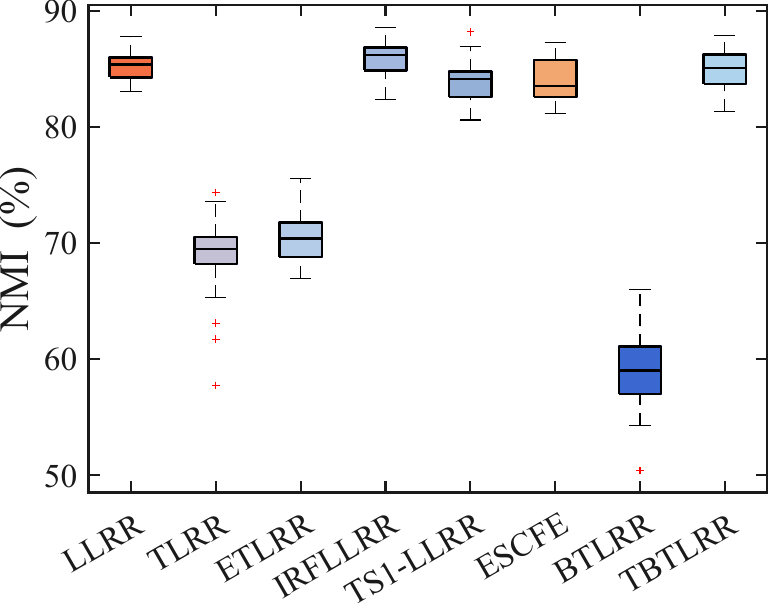}
}\hspace{0mm}
\subfigure[Extended YaleB (NMI)]{
    \label{h}
    \centering
    \includegraphics[width=3.9cm]{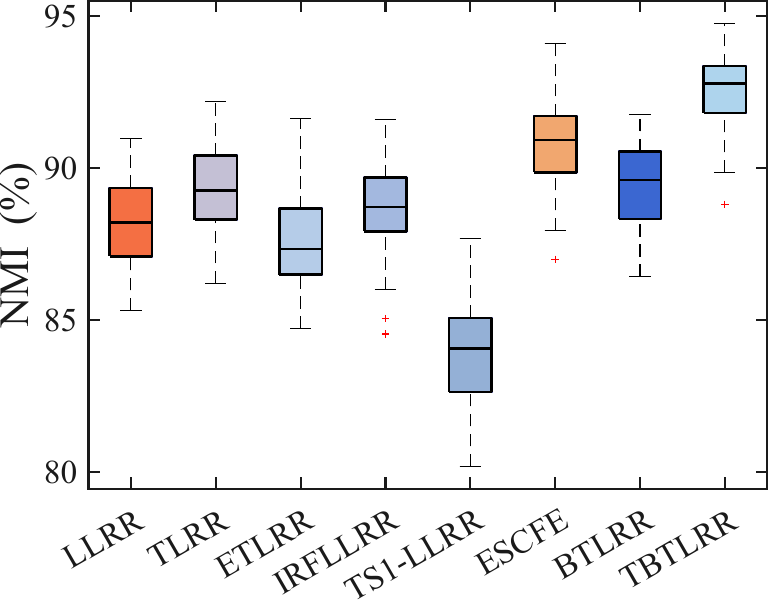}
}\hspace{0mm}
\subfigure[UCSD (NMI)]{
    \label{g}
    \centering
    \includegraphics[width=3.9cm]{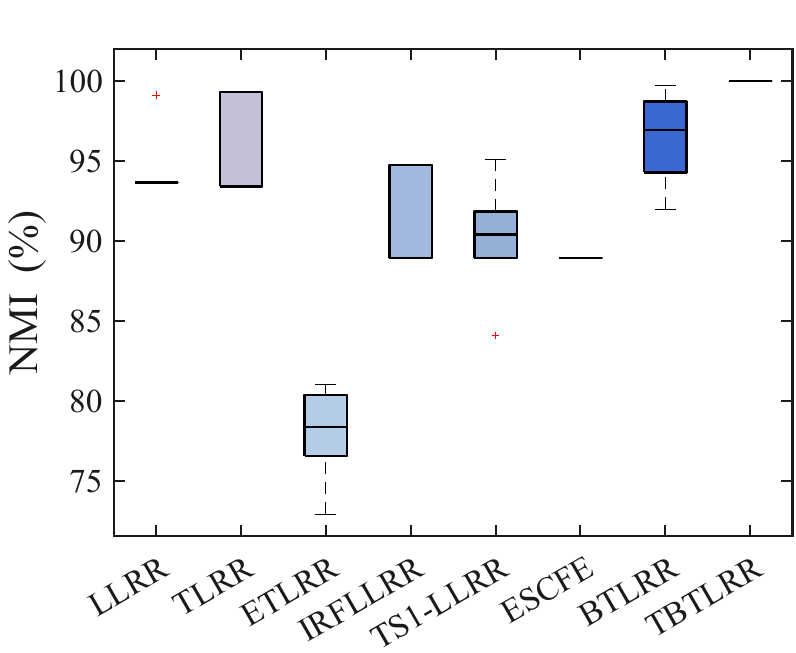}
}\hspace{0mm}
\caption{Model stability comparisons of all compared methods on four real-world datasets, where (a)-(d) are the ACC results and (e)-(h) are the NMI results.}
\label{stability}
\end{figure*}

\begin{figure*}[t]
\hspace{-0.5cm}
\centering
\subfigure[ORL]{
    \label{a}
    \centering
    \includegraphics[width=4cm]{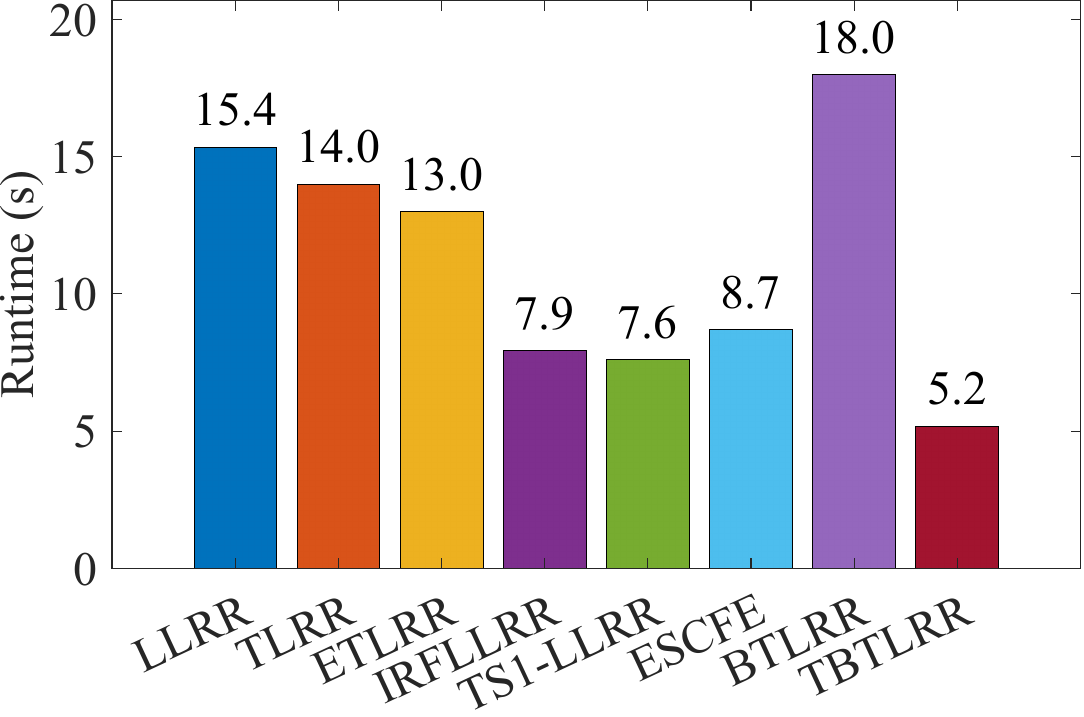}
}\hspace{0mm}
\subfigure[Umist]{
    \label{b}
    \centering
    \includegraphics[width=4cm]{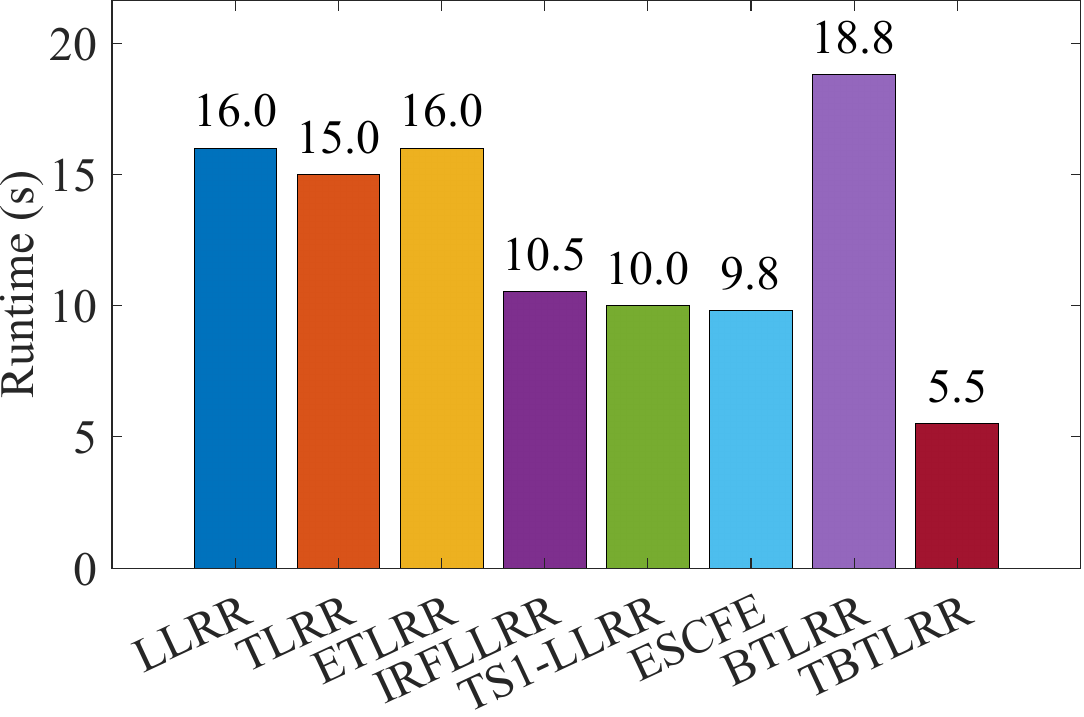}
}\hspace{0mm}
\subfigure[Extended YaleB]{
    \label{c}
    \centering
    \includegraphics[width=4cm]{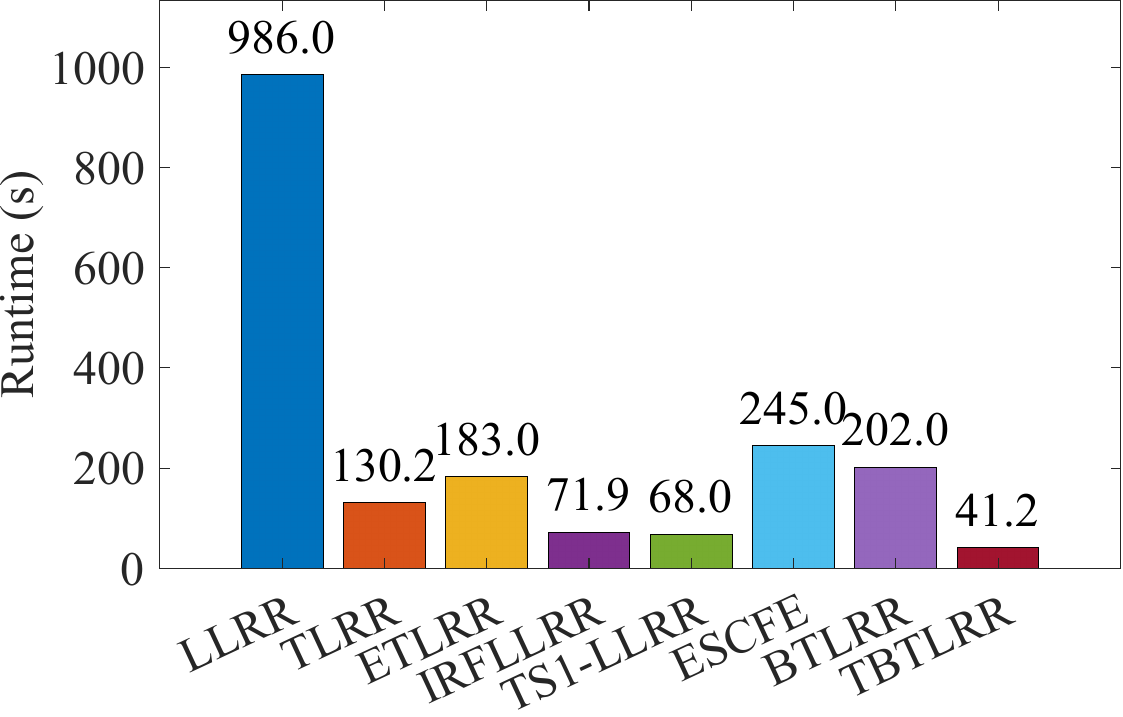}
}\hspace{0mm}
\subfigure[UCSD]{
    \label{c}
    \centering
    \includegraphics[width=4cm]{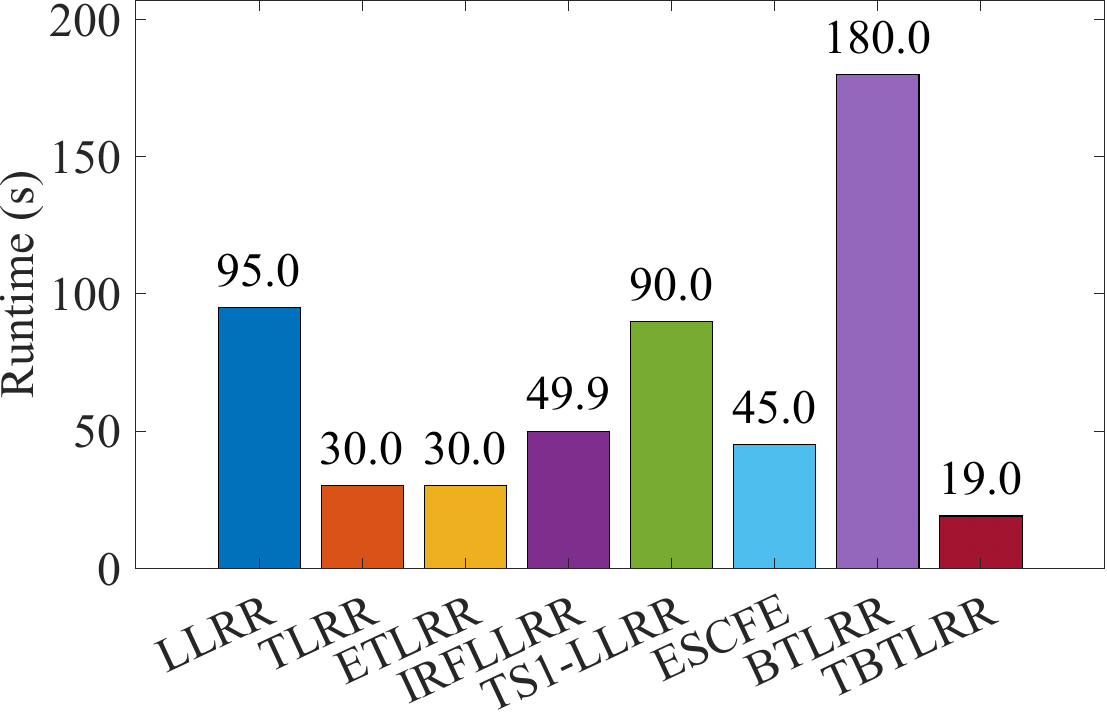}
}\hspace{0mm}

\hspace{-0.5cm}
\subfigure[Notting-Hill]{
    \label{e}
    \centering
    \includegraphics[width=4cm]{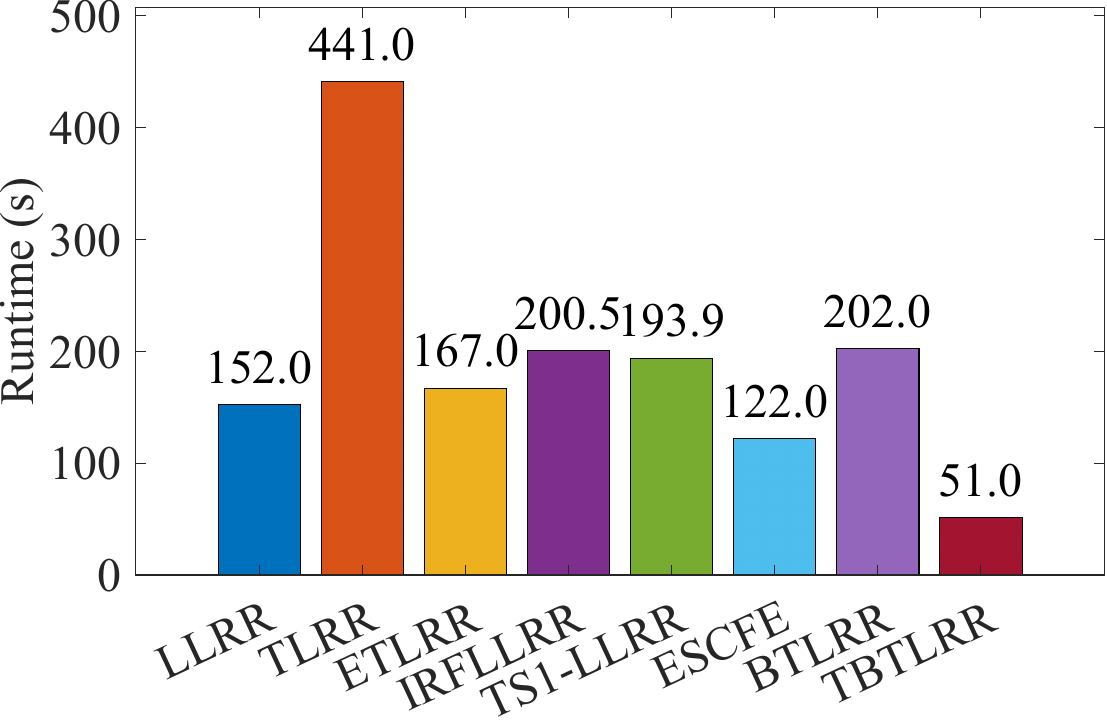}
}\hspace{0mm}
\subfigure[MSTAR]{
    \label{f}
    \centering
    \includegraphics[width=4cm]{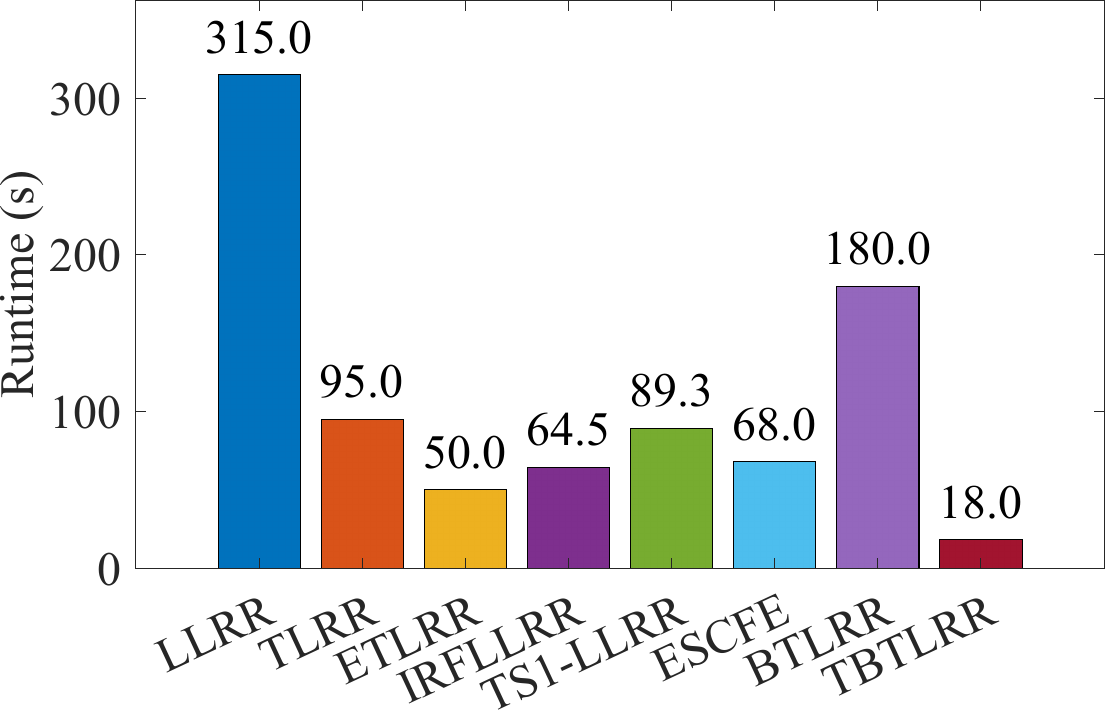}
}\hspace{0mm}
\subfigure[GTSRB]{
    \label{g}
    \centering
    \includegraphics[width=4cm]{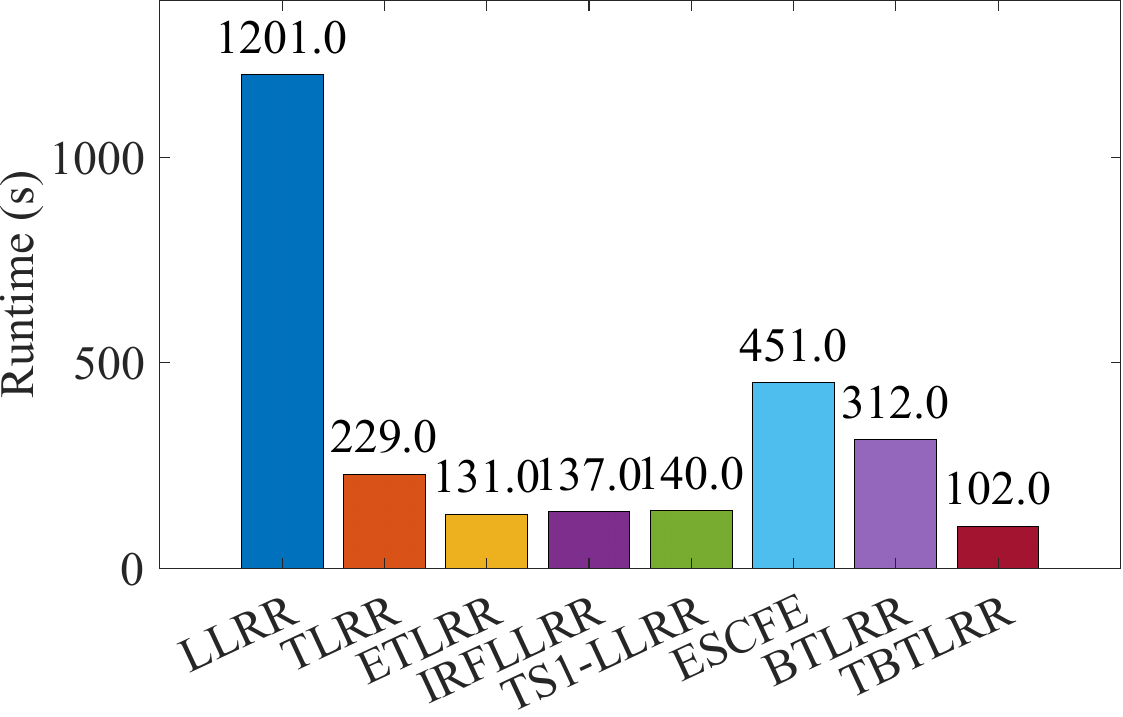}
}\hspace{0mm}
\subfigure[MNIST]{
    \label{g}
    \centering
    \includegraphics[width=4cm]{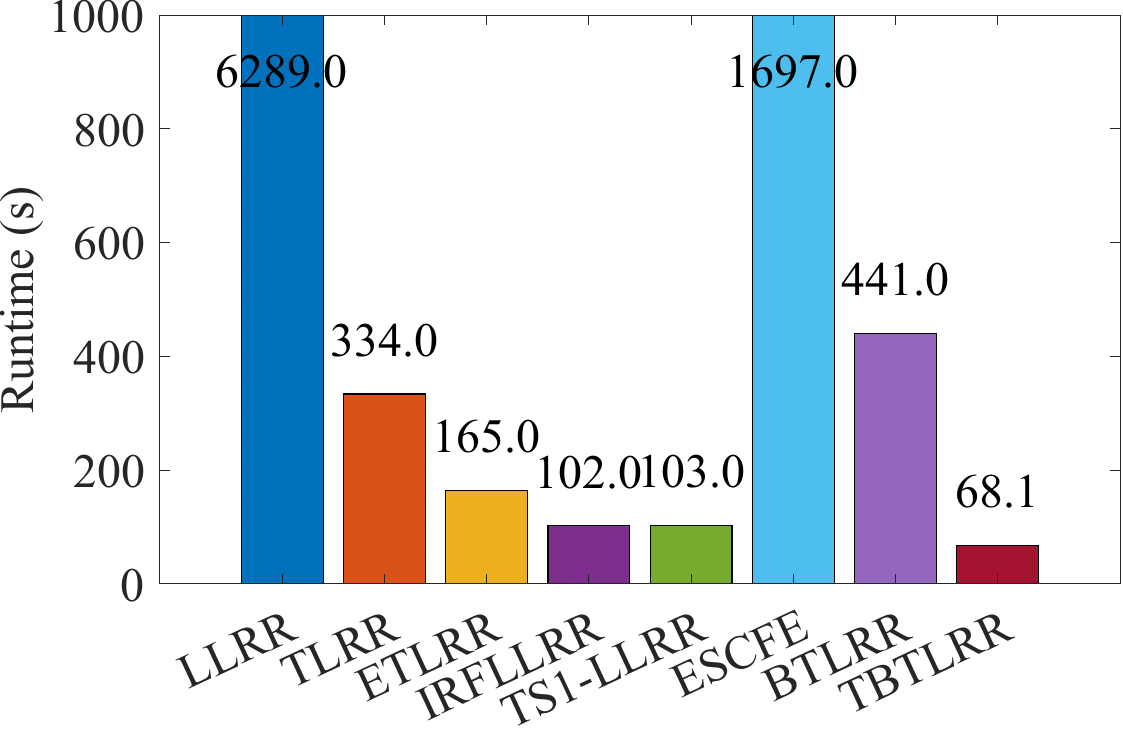}
}\hspace{0mm}

\caption{Runtime (seconds) of compared methods on different datasets. }
\label{TIME}
\end{figure*}

\subsection{Ablation Studies}

The proposed TBTLRR integrates the $\ell_{1/2}$-norm and Frobenius norm  within a unified framework. To verify their effectiveness, four comparative variants are designed, including
\begin{itemize}
  \item Case I: TBTLRR without $\|\mathcal{E}\|_{1/2}$ and $\|\mathcal{N}\|_{F}^{2}$
  \item Case II: TBTLRR without $\|\mathcal{E}\|_{1/2}$
  \item Case III: TBTLRR without $\|\mathcal{N}\|_{F}^{2}$
  \item Case IV: TBTLRR 
\end{itemize}

As shown in Table \ref{ablation}, on the ORL dataset, introducing the $\ell_{1/2}$-norm improves ACC from 67.56\% to 72.50\% and NMI from 82.17\% to 85.04\%. On the Umist dataset, retaining the Frobenius term is critical, as removing it degrades ACC from 79.90\% to 75.71\% and NMI from 85.73\% to 81.20\%. On the Extended YaleB dataset, combining both regularizers yields the best results. Overall, the two regularizers contribute in complementary ways while yielding consistent overall performance improvements.


\subsection{Discussion}\label{Discussion}

\subsubsection{Parameter Analysis}

Fig. \ref{Para} shows that the optimal configurations of $\lambda$ and $\beta$ differ across datasets. On the Umist and Extended YaleB datasets, increasing $\beta$ improves both ACC and NMI, highlighting its contribution to regularization. On the ORL and UCSD datasets, the interaction between $\lambda$ and $\beta$ becomes more significant, with an appropriately selected $\lambda$ enhancing clustering performance. Proper parameter selection helps preserve data structure and mitigate noise, leading to improved clustering results. Thus, the choice of regularization parameters should be tailored to the dataset's characteristics.

\subsubsection{Model Stability}

Fig. \ref{stability} presents the boxplots of 50 clustering results. As shown in the figure, our proposed TBTLRR demonstrates superior average performance across all datasets. When other methods exhibit similar performance to TBTLRR, the boxplot corresponding to TBTLRR is noticeably narrower, indicating that it has more stable performance in the experiments.

\subsubsection{Complexity Comparison}
Fig. \ref{TIME} compares the runtime of different methods across eight datasets. It is found that our proposed TBTLRR achieves the fastest runtime on all datasets. Note that TLLRR is the slowest among all methods because it does not employ the subspace projection technique used by most competitors (except ESCFE). Moreover, TBTLRR adopts a data-adaptive transformation, avoiding the additional complexity introduced by the FFT used in conventional TLRR, and thus attains superior computational efficiency.

\section{Conclusion}\label{Conclusion}

This paper proposes data-adaptive transformed bilateral tensor LRR called TBTLRR that introduces adaptive unitary transformations to effectively capture the global low-rank structures, making the model more flexible and efficient for various data types. To enhance robustness against noise, TBTLRR integrates a joint regularization strategy using the $\ell_{1/2}$-norm and Frobenius norm. Additionally, a diagonal-ratio-based weighted fusion strategy is introduced to assign appropriate weights to tensor slices, thereby improving the stability of the affinity matrix and reducing the impact of noise on the final clustering results. Extensive studies validate the effectiveness and robustness on benchmark datasets.

In future work, we are interested in exploring deep unfolding networks to endow this method with learnable priors and automatic parameter tuning.

\bibliographystyle{IEEEtran}
\bibliography{mybib}

\end{document}